\documentclass[lettersize,journal]{IEEEtran}
\usepackage{amsmath,amsfonts}
\usepackage{algorithmic}
\usepackage{algorithm}
\usepackage{array}
\usepackage[caption=false,font=normalsize,labelfont=sf,textfont=sf]{subfig}
\usepackage{textcomp}
\usepackage{stfloats}
\usepackage{url}
\usepackage{verbatim}
\usepackage{graphicx}
\usepackage{cite}
\usepackage[table]{xcolor}
\usepackage{booktabs}
\usepackage{multirow}
\usepackage{orcidlink}

\hypersetup{
colorlinks=true,
linkcolor=black,
citecolor=black
}

\begin{document}

\title{Mug-STAN: Adapting Image-Language Pretrained Models for General Video Understanding}

\author{Ruyang Liu$^{\orcidlink{0009-0002-6664-6763}}$, Jingjia Huang$^{\orcidlink{0000-0002-0834-3265}}$, Wei Gao$^{\orcidlink{0000-0001-7429-5495}}$, Thomas H. Li$^{\orcidlink{0000-0001-6123-1265}}$, Ge Li$^{\orcidlink{0000-0003-4079-3968}}$
%,~\IEEEmembership{Staff,~IEEE,}
        % <-this % stops a space
\thanks{Ruyang Liu, Wei Gao, Thomas H. Li, and Ge Li are with the School of Electronic and Computer Engineering, Peking University Shenzhen Graduate School. Ruyang Liu is also with the Peng Cheng Laboratory. 
E-mail: \{ruyang@stu, gaowei262@, geli@ece., thomas@\}pku.edu.cn}% <-this % stops a space
\thanks{Jingjia Huang is with ByteDance Inc. E-mail: huangjingjia@bytedance.com}
\thanks{Corresponding author: Gao Wei.}}

\markboth{Journal of \LaTeX\ Class Files,~Vol.~14, No.~8, May~2023}%
{Shell \MakeLowercase{\textit{et al.}}: A Sample Article Using IEEEtran.cls for IEEE Journals}

%\IEEEpubidadjcol

\maketitle

\begin{abstract}
    Large-scale image-language pretrained models, \textit{e.g.}, CLIP, have demonstrated remarkable proficiency in acquiring general multi-modal knowledge through web-scale image-text data. Despite the impressive performance of image-language models on various image tasks, how to effectively expand them on general video understanding remains an area of ongoing exploration. 
    In this paper, we investigate the image-to-video transferring from the perspective of the model and the data, unveiling two key obstacles impeding the adaptation of image-language models: non-generalizable temporal modeling and partially misaligned video-text data.
    %In this paper, we unveil that the lack of generalizability in temporal modeling is the key obstacle impeding the adaptation of image-language models to the video domain. Existing temporal modeling methods often lack universality across different video tasks and are ill-suited for handling partially misaligned video-text data.
    %In this paper, we uncover two pivotal factors hindering the adaptation of image-language models to the video domain: non-generalizable temporal modeling and partially misaligned video-text data, whereas existing methods usually concentrate on one specific problem and single video task. 
    To address these challenges, we propose \textit{S}patial-\textit{T}emporal \textit{A}uxiliary \textit{N}etwork with \textit{Mu}tual-\textit{g}uided alignment module (Mug-STAN) – a simple yet effective framework extending image-text model to diverse video tasks and video-text data. Specifically, STAN adopts a branch structure with decomposed spatial-temporal modules to enable generalizable temporal modeling, while Mug suppresses misalignment by introducing token-wise feature aggregation of either modality from the other. Extensive experimental results verify Mug-STAN significantly improves adaptation of language-image pretrained models such as CLIP and CoCa at both video-text post-pretraining and finetuning stages. With our solution, state-of-the-art zero-shot and finetuning results on various downstream datasets, including MSR-VTT, DiDeMo, LSMDC, Kinetics-400, Something-Something-2, HMDB-51, UCF-101, and AVA, are achieved. Moreover, by integrating pretrained Mug-STAN with the emerging multimodal dialogue model, we can realize zero-shot video chatting. Codes are available at \href{https://github.com/farewellthree/STAN}{https://github.com/farewellthree/STAN}
\end{abstract}


\section{Introduction}
\IEEEPARstart{I}{n} the past three years, the computer vision community has witnessed the remarkable success of web-scale pretrained image-language models, such as CLIP \cite{radford2021learning}, CoCa \cite{yu2022coca}, and BEiTv3 \cite{wang2023image}. However, the development of fundamental video-language models is challenging, due to the high cost of computation resources needed for pretraining and the limited availability of data in terms of scale, quality, and diversity. Rather than focusing on developing video-language pretrained models \cite{miech2019howto100m,huang2023clover}, an alternative and promising approach is to transfer the abundant knowledge in image-language pretrained models to the video domain, which has garnered increasing attention in recent years \cite{xue2022clip,luo2022clip4clip,liu2023revisiting,buch2022revisiting,ni2022expanding,pan2022st}.

%Extending pretrained 2D image models to the realm of videos has been extensively investigated within the field of video learning \cite{carreira2017quo,bertasius2021space}. One readily conceivable avenue of research pertains to temporal modeling, with the goal of endowing 2D models with the capacity to capture temporal dependencies among video frames. In theory, a properly designed temporal modeling module should be capable of generalization across various video tasks and datasets. However, a notable challenge arises from the fact that most existing temporal modeling methods in image-language models are tailored to specific video tasks and exhibit subpar performance when confronted with misaligned video-text data.

\begin{figure*}[h]
    \centering
    \setlength{\abovecaptionskip}{-0.1 cm}
    \includegraphics[width=1\linewidth]{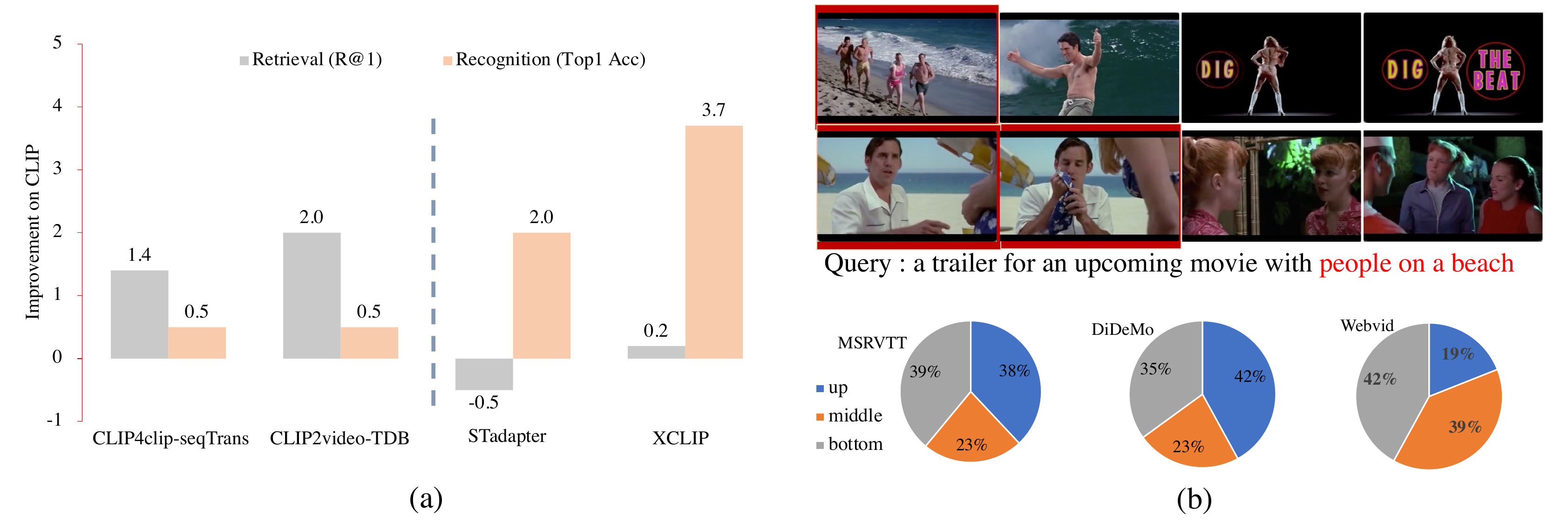}
    \caption{The two issues in image-to-video transfer for vision-language models. (a) Generalizability: We illustrate CLIP-based temporal modules struggle to generalize across different video tasks. We present the performance of various models concerning the baseline, which is based on \textit{CLIP} with mean pooling. The models include the text-video retrieval models CLIP4clip-seqTrans \cite{luo2022clip4clip} and CLIP2video-TDB \cite{fang2021clip2video}, as well as video recognition models STadapter \cite{pan2022st} and XCLIP \cite{ni2022expanding}. Evaluation is based on Recall@1 for MSRVTT \cite{xu2016msr} and Top-1 accuracy for Kinetics-400 \cite{kay2017kinetics}. (b) Partial Misalignment: Above, we showcase a misaligned training sample in MSRVTT, where only ``people on a beach" and the $1^{st}, 5^{th} and \ 6^{th}$ frames are aligned to each other. Below, we quantitatively assess the extent of partial misalignment in video-text datasets, including MSRVTT, DiDeMo \cite{anne2017localizing}, and WebVid2.5M \cite{bain2021frozen}. The degree of alignment progressively deteriorates from ``up'' to ``bottom''.}
    \label{intro1}
\end{figure*}

The extension of pretrained 2D image models to the realm of videos has been extensively explored within the field of video learning \cite{carreira2017quo, bertasius2021space}. The central challenge lies in the modality disparity between images and videos. Specifically, videos inherently contain unique temporal information, and video-text data is generally more complex and noisy when compared to image-text data. Consequently, our investigation, built upon existing temporal modeling methods and various video-language datasets, has revealed two often overlooked points. As depicted in Fig. \ref{intro1}, we have found that current efforts in temporal modeling are predominantly confined to either video-language tasks \cite{luo2022clip4clip, fang2021clip2video, liu2022ts2, xue2022clip} or video-specific tasks \cite{ni2022expanding, pan2022st, buch2022revisiting}, resulting in reduced efficiency when applied to a different category of video task. Meanwhile, our observation indicates that video-text paired training samples typically suffer from partial misalignment in both pretraining and downstream datasets.

\begin{figure}
    \centering
    \setlength{\abovecaptionskip}{-0.05 cm}
    \includegraphics[width=1\columnwidth]{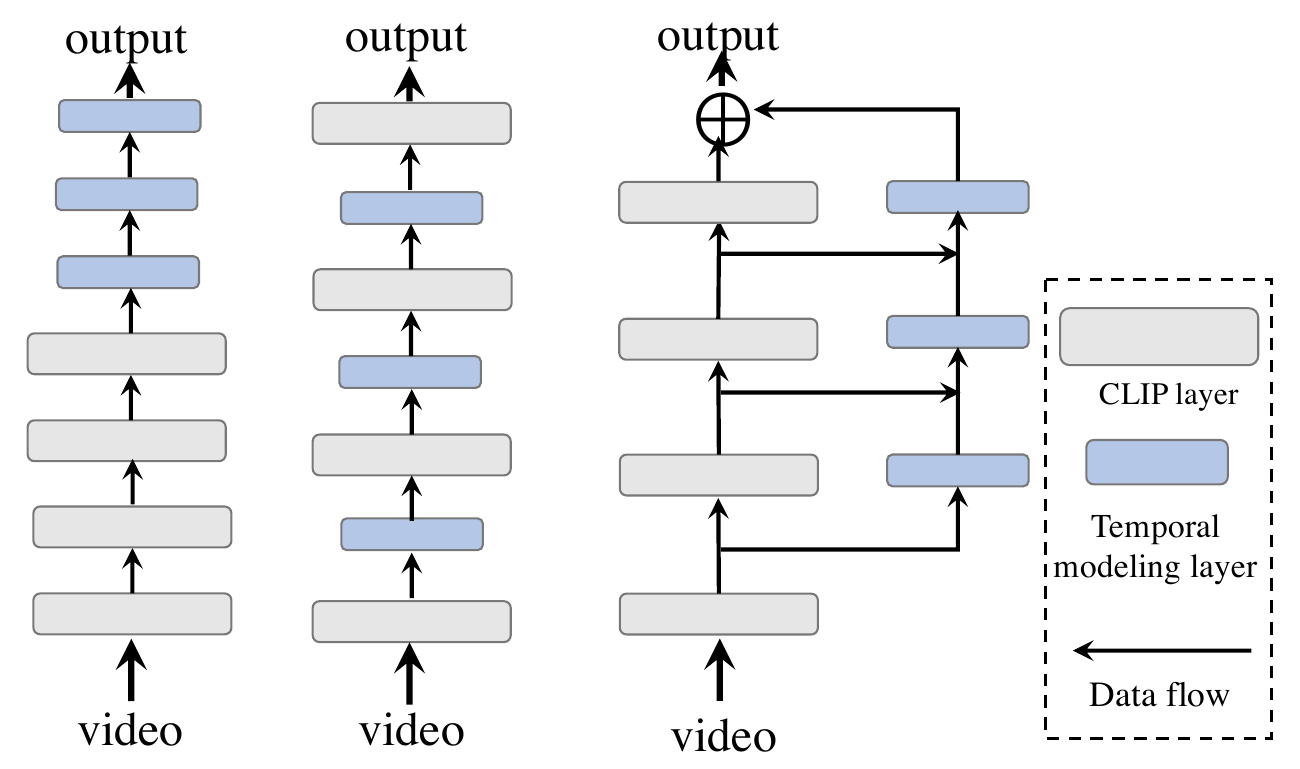}
    \caption{Different structures of temporal modeling: posterior structure (left), intermediate structure (middle), and our branch structure (right).}
    \label{intro2}
    \vspace{-0.7em}
\end{figure}

\begin{figure}
    \centering
    \setlength{\abovecaptionskip}{-0.05 cm}
    \includegraphics[width=1\columnwidth]{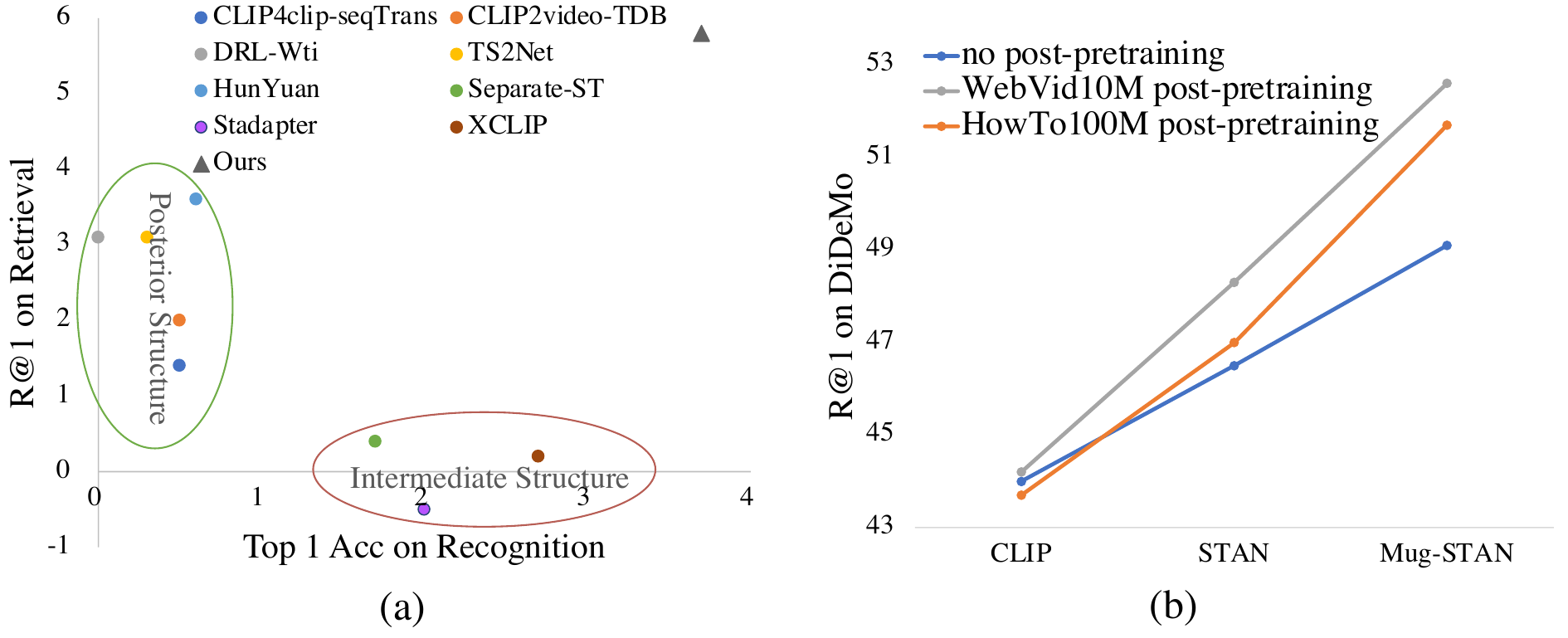}
    \caption{(a) Performance comparison of various methods on both text-to-video retrieval and video recognition. Evaluation is based on Recall@1 for MSRVTT \cite{xu2016msr} and Top-1 accuracy for Kinetics-400 \cite{kay2017kinetics}. The methods are clustered into posterior structure, intermediate structure, and our branch structure. (b) Performance comparison of post-pretraining on different models. We report the finetuned result of Recall@1 on DiDemo text-video retrieval. Based on \textit{CLIP}, effective temporal modeling (STAN) and partial-misalignment suppression (Mug) respectively bring noticeable improvements.
    }
    \label{intro3}
    \vspace{-0.7em}
\end{figure}

To gain a deep insight into the first issue, we further dive into the structures of existing CLIP-based temporal modules. We find current efforts can be roughly categorized into posterior structure based methods and intermediate structure based methods as shown in Fig. \ref{intro2}. Posterior structure based methods \cite{luo2022clip4clip, fang2021clip2video, liu2022ts2, wang2022disentangled, jiang2022tencent} adopt a late modeling strategy, utilizing CLIP as a feature extractor and applying temporal modeling to embeddings independently extracted from different frames. Built upon the highly semantic embeddings, this structure, while beneficial for preserving well-aligned visual-language representations, falls short in capturing the low-level spatial-temporal visual patterns among frames, which are essential for video understanding. 
As a result, methods based on posterior structures tend to exhibit marginal performance improvements, a trend that becomes particularly pronounced in action recognition tasks where low-level spatial-temporal visual patterns are crucial.
Unlike posterior structure based methods, intermediate structure based methods \cite{ni2022expanding, pan2022st, bertasius2021space} equip CLIP with temporal modeling capability by integrating temporal modeling modules between CLIP layers, which sees significant improvements in the video recognition task. Nevertheless, we have observed that incorporating additional modules inside CLIP would impact the pretrained high-level semantic knowledge in the model, leading to trivial or even negative impacts on the text-video retrieval task. These statistical patterns are more pronounced in Fig. \ref{intro3}, where both the posterior structure and intermediate structure excel only in their respective tasks.

In contrast to the extensive research on temporal modeling, another critical issue has received limited attention: video-text paired training samples generally exhibit partial misalignment.
Partial misalignment refers to the situation in which the aligned information between a video and its corresponding text is distributed only across specific frames and phrases, while other components of the video/text are noisy which hinders precise vision-language alignment and strong image-to-video adaptation. Fig. \ref{intro1}(b) shows a case of partial misalignment, where only the phrase ``people on a beach'' and the red-marked frames are semantically aligned. Due to the complexity and redundancy of video content, such cases occur much more frequently in video-text than in image-text data.
Moreover, the situation is even more severe in video pretraining datasets, which are constructed using instructional videos and noisy narrations \cite{miech2019howto100m, xue2022hdvila, zellers2021merlot}. 
To quantitatively assess the partial misalignment present in video datasets, we have selected and analyzed two downstream datasets (MSR-VTT \cite{xu2016msr} and DiDeMo \cite{anne2017localizing}) and one pretraining dataset (WebVid2.5m \cite{bain2021frozen}). Specifically,  we employ CLIP-ViT-L/14 \cite{radford2021learning} to measure misalignment, utilizing dot-product similarity followed by sigmoid to compute the correlation between text and each frame. A frame is considered aligned with the text if the probability exceeds 0.5. Then, we categorize the video-text alignment degree into three levels: (1)up when more than 2/3 frames are aligned with the text. (2)bottom when less than 1/3 frames are aligned with the text. (3)middle in between the two. As revealed in Fig. \ref{intro1}(b), in all three datasets, more than half of the video-text pairs suffer from partial misalignment (middle and bottom), even if these datasets are widely recognized for their high quality in video-text tasks. 

Partial misalignment, together with the temporal modeling, has raised a subsequent challenge: post-pretraining\footnote{Further pretraining on relatively large scale video-text corpora based on pretrained image models for downstream video tasks is termed as post-pretraining. Finetuning means directly tuning for adapting image-text models on downstream video datasets.} image-language models on large-scale video-language datasets shows very limited gains. As depicted in Fig. \ref{intro3}(b), we can observe that CLIP, after being post-pretrained on either WebVid10M or HowTo100M, does not significantly outperform the baseline without post-pretraining.

%除了复制，还要提到数据，现有工作集中在howto上

From the aforementioned analysis, we conclude two key factors for extending image-language pretrained models to the video domain: (1) Effective temporal modeling while taking advantage of knowledge in different levels of representation. (2) Suppressing the partial misalignment during training on video-text data. To this end, we propose \textbf{S}patial-\textbf{T}emporal \textbf{A}uxiliary \textbf{N}etwork with \textbf{Mu}tual-\textbf{g}uided alignment module (Mug-STAN) - a plug-and-use framework adapting image-language models to general video tasks, where STAN introduces effective temporal modeling and Mug mitigates partial misalignment during training. In Fig. \ref{intro2} and \ref{intro3}(a), it is noticeable that temporal modeling structure in STAN exhibits strong performance in both retrieval tasks and recognition tasks. In Fig. \ref{intro3}(b), we can see that STAN and Mug contribute significantly to the effectiveness of post-pretraining respectively, where Mug excels particularly well on the noisy HowTo100M dataset. 

Specifically, rather than posterior or intermediate structure, our proposed STAN introduces a distinctive branch structure located outside the visual backbone , featuring multiple levels of input, as shown in Fig. \ref{intro2}. This novel structure enables STAN to enrich the features of video frames with spatial-temporal contexts, leveraging different output levels of image-text model, while preserving the forward-propagation of source model. Thereby, it can effectively utilizes both high-level and low-level knowledge from the pretrained model simultaneously, making it adaptable to various downstream video tasks. STAN comprises multiple layers with a spatial-temporal separated design. Each layer conducts spatial-temporal modeling by alternately stacking two distinct modules: an intra-frame module and a cross-frame module. This approach allows the layer to enhance model performance by reusing pretrained parameters from image-text pretrained models to initialize the intra-frame spatial modules. Meanwhile, Mug is constructed using a parameter-free, token-wise interaction modeling mechanism with negligible computational cost, which can be easily plugged into existing state-of-the-arts. Given a video-text pair, we can get its frame-wise feature sequence and text feature sequence, respectively. To realize the mutual-guided alignment, we first perform the frame-token interaction to obtain the frame-specific text embedding for each frame and token-specific video embedding for each token. Then, for each modality, we attain its final global embedding through guidance from the other modality. At last, the pair of mutually guided representations are employed in contrastive learning during post-pretraining or finetuning. In this way, we can capture and align the relevant parts of video and text, freeing the adaptation of image-text pretrained models from the video-text partial misalignment problem.

Through extensive experiments, we have demonstrated the impressive performance of our proposed Mug-STAN. Specifically, we have implemented Mug-STAN on two well-known image-language models, CLIP and CoCa. Furthermore, we have adopted a fresh perspective on post-training by evaluating our model on datasets with varying levels of noise, such as WebVid10M and HowTo100M. The comprehensive results highlight the efficacy of Mug-STAN not only in the finetuning but also in post-pretraining. Remarkably, we achieve state-of-the-art results in both zero-shot and finetuning settings across a diverse range of video tasks, including text-video retrieval, video action recognition, and video detection. Moreover, given the current popularity of multimodal dialogue systems, we have also plugged the pretrained Mug-STAN on LLaVa \cite{liu2023visual}, achieving the capability of zero-shot video chatting without any instruction tuning.

The main contributions of this paper are:
\begin{itemize}
\item We present an in-depth analysis of the factors that impede the adaptation of image-language models to video domains. By revisiting the temporal modeling on CLIP in current research and carefully examining video-text datasets, we identify non-generalizable temporal modeling and partially misaligned video-text data as the primary culprits affecting the performance.
\item We propose \textbf{S}patial-\textbf{T}emporal \textbf{A}uxiliary \textbf{N}etwork with \textbf{Mu}tual-\textbf{g}uided alignment module (Mug-STAN) - a simple but strong framework that extends image-text pretrained models to general video tasks. In Mug-STAN, we leverage the novel branch structure of STAN for effective temporal modeling, enabling temporal learning that incorporates spatial-temporal contexts at various levels. Additionally, Mug plays a crucial role in noise suppression and encourages the contribution of well-aligned parts to achieve robust video-language alignment.
\item We conduct comprehensive experiments under various settings to evaluate the effectiveness of Mug-STAN. The numerous results demonstrate that Mug-STAN achieves state-of-the-art zero-shot and finetuning results on a wide range of video datasets and tasks, as well as the capability of zero-shot video dialogue.
\end{itemize}

\section{Related Work}
\subsection{Image-Language PreTraining}
Image-Language pre-training has been drawing increasing attention from researchers in the computer vision community \cite{zellers2021merlot, huang2021seeing, huang2020pixel, xue2021probing}. Recently, contrastive language-image pretraining on web-scale data \cite{radford2021learning, yu2022coca, wang2023image, jia2021scaling, yuan2021florence} has experienced significant success, primarily due to its outstanding performance when applied to various downstream tasks. One of the most renowned works is CLIP \cite{radford2021learning}, which has demonstrated surprising capabilities in zero-shot recognition and domain generalization \cite{zhou2022learning, patashnik2021styleclip}. 
The wealth of knowledge contained within these image-language pretrained models holds a promising future for their adaptation to video tasks. Thankfully, our Mug-STAN can be implemented on these image-language models in a plug-and-play manner, leading to substantial performance improvements in various video tasks. It's worth noting that recent advancements in multimodal understanding have been largely propelled by the fusion of image-based vision models with LLMs, such as Flamingo \cite{alayrac2022flamingo}, BLIP-2 \cite{li2023blip}, and LLaVA \cite{liu2023visual}, fortunately, these multimodal dialogue models generally employ CLIP-L/14 as the visual encoder. Consequently, our Mug-STAN can be seamlessly implemented on these models to achieve zero-shot video chatting.

\subsection{Video-Language Pretraining}
As a subset of vision-language pretraining, video-language pretraining has also been the subject of numerous explorations in recent years, such as Violet \cite{fu2021violet}, clipBert \cite{lei2021less}, Frozen \cite{bain2021frozen}, BridgeFormer \cite{ge2022bridging}, and Clover \cite{huang2023clover}. In video-language pretraining, models typically initialize the video encoder and text encoder with separately pre-trained weights \cite{arnab2021vivit, bertasius2021space, liu2022video, devlin2018bert}, and then use multiple pretraining targets to achieve cross-modal alignment and multimodal learning, such as contrastive learning, masked language modeling, and video-text matching. However, video-language pretrained models face difficulties in simultaneously handling temporal modeling and modality alignment due to the challenges posed by unaligned initialization. In contrast, image-text pretrained models inherently possess extensive knowledge as a result of the vast diversity and scale of image-text data they are trained on. As a result, when finetuned on downstream video-language datasets, we have observed significant advantages of image-text pretrained models over video-language pretrained models, even if the former have not been pretrained on video datasets.

%Similar to our research, CLIP-ViP \cite{xue2022clip} is among the few studies that delve into the realm of video post-pretraining. However, CLIP-ViP relies on large-scale data and the annotation from an additional captioner for its post-pretraining process. In contrast, our work demonstrates that with an appropriate method, post-pretraining can yield superior results on both smaller datasets (Webvid10M) and noisy datasets (HowTo100M) without requiring extra frame-wise annotation. In addition, there are also some studies that explore the realm of pretraining under noisy and misaligned video-text data \cite{miech2020end, han2022temporal, zeng2023learning}. For instance, Miech \textit{et al} \cite{miech2020end} presented a MIL-NCE loss for noisy video-narration pretraining, where the content of text may arise in the previous or the following clips. Han \textit{et al} \cite{han2022temporal} proposed a Temporal Alignment Network to manage denoising and aligning simultaneously. Unlike these works, our method does not primarily concentrate on datasets filled with completely misaligned or unrelated video-text pairs (ASR) \cite{miech2019howto100m, zellers2021merlot}. Instead, we specifically address the issue of partial misalignment, which remains prevalent even in relatively high-quality datasets (as depicted in Fig. \ref{intro2}(b)).

Similar to our research, CLIP-ViP \cite{xue2022clip} is among the few studies that delve into the realm of video post-pretraining. However, CLIP-ViP relies on large-scale data and the annotation from an additional captioner for its post-pretraining process. In contrast, our work demonstrates that with an appropriate method, post-pretraining can yield superior results on both smaller datasets (Webvid10M) and noisy datasets (HowTo100M) without requiring extra frame-wise annotation. In addition, several studies have also ventured into the domain of pretraining under noisy and misaligned video-text data \cite{miech2020end, zeng2023learning, han2022temporal}. Miech et al. \cite{miech2020end} and Han et al. \cite{han2022temporal} introduced the MIL-NCE loss and the Temporal Alignment Network, respectively, for noisy video-narration pretraining. Compared to these works, our paper differs in three aspects: (1)Setting. The previous works primarily focus on the datasets filled with completely misaligned video-text pairs and ASR captions (\textit{e.g., Howto100M}), while our focus lies on the issue of partial misalignment, which is a more general problem and can even occur in relatively high-quality datasets, as depicted in Fig. \ref{intro2}(b). (2) Method. \cite{han2022temporal} employs the black-box network to learn the similarity between video and text, while we propose a parameter-free video-text mutual-guided module to identify and filter out the unrelated parts from video and text. (3) Results. In experiments, we convey much better results than those works under the same setting.

\subsection{Image-Language Pretrained Models For Video Tasks}
In contrast to further post-pretraining, the majority of current studies primarily concentrate on the direct fine-tuning of image-text models for video tasks. An intuitive direction is temporal modeling \cite{luo2022clip4clip, fang2021clip2video, liu2022ts2, xue2022clip, ni2022expanding, pan2022st, buch2022revisiting, gao2021clip2tv, zhang2023temporal, liu2023visual}, as the image model cannot capture temporal information. 
In video-language tasks, such as text-video retrieval, most adaptation models tend to utilize posterior-based structures to handle temporal aspects , \textit{e.g.,}, the sequential transformer in \cite{luo2022clip4clip}, the temporal difference block in \cite{fang2021clip2video}, and token selection module in \cite{liu2022ts2}. Despite the advancements achieved by these methods, the temporal modeling they provide is restricted to high-level embeddings and lacks effectiveness, as illustrated in Fig. \ref{intro1}(a). In video-only tasks such as action recognition, the mainstream expansion of CLIP for temporal modeling is to utilize the intermediate structure. For instance, Ni \textit{et al} \cite{ni2022expanding} developed a message token mechanism to pass messages among different frames.  Pan \textit{et al} \cite{pan2022st} inserted the 3D convolution adapter inside the transformer to activate temporal modeling. Besides temporal modeling, there are also other efforts focused on adapting image-language models for video tasks from different perspectives. For example, \cite{ni2022expanding, ju2022prompting} explored the prompt modeling, while \cite{wang2022disentangled, liu2022ts2, hu2023cross, liu2021aligning} improved the ways of cross-modal interaction. However, most of the aforementioned methods tend to perform worse when transferred to another video task, whereas our model performs well across various video tasks.

\section{Method}
In this section, we will elaborate on our proposed strong and flexible Mug-STAN for adapting image-language models to general video tasks.

\subsection{Motivation} \label{method1}
\begin{figure*}[h]
    \centering
    \setlength{\abovecaptionskip}{0.1 cm}
    \includegraphics[width=1\linewidth]{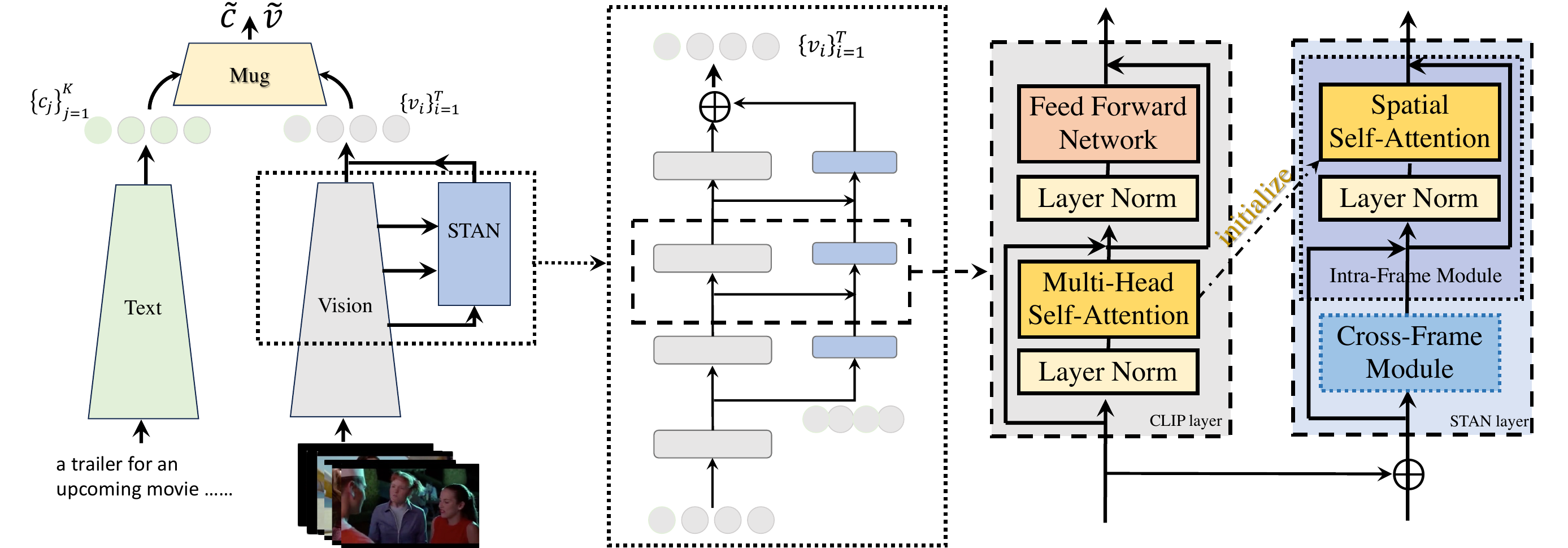}
    \caption{(left) The overall architecture of our proposed method, including the text and visual encoders, the temporal modeling module (STAN), and the cross-modal interaction module (Mug). (middle) Schematic diagram of feature forward propagation in and between pretrained visual encoder and STAN. (right) Details of the internal structure of the STAN spatial-temporal module.}
    \label{img_method1}
    \vspace{-1em}
\end{figure*}

Large-scale image-language models, such as CLIP and CoCa, which undergo pretraining on hundreds of millions to billions of image-text pairs, typically comprise two encoders as fundamental components.
Each encoder is responsible for encoding one modality to facilitate cross-modal alignment. As we ascend through the layers of the visual transformer \cite{dosovitskiy2020image}, the model gradually learns visual patterns at different levels of abstraction \cite{yuan2021tokens}. Eventually, the visual encoder produces high-level visual embeddings that are semantically aligned with the corresponding embeddings in the text modality. Formally, as illustrated in Fig. \ref{img_method1}(left), given a video clip with $T$ frames and a text description with $K$ tokens, we feed them into a standard image-text pretrained visual encoder and text encoder, treating each frame as an individual image. This process generates frame-wise video representations denoted as $V$, and token-wise text representations denoted as $C$:
\begin{equation}
        V = \{ v_i \}_{i=1}^{T} \in \mathbb{R}^{T \times D}, \quad
        C = \{ c_j \}_{j=1}^{K} \in \mathbb{R}^{K \times D}
        \label{eq_represent}
 \end{equation}
where $D$ is the feature dimension. Note that $v_i$ can be obtained from either the CLS token \cite{radford2021learning, yu2022coca} or the average of all patch tokens \cite{yuan2021florence} of each frame. Then, frame-wise video representations $\{ v_i \}_{i=1}^{T}$ are averaged as the global video embedding $v$ and the CLS token embedding is chosen from $C$ as the global text representation $c$, where $v$ and $c$ are employed for cross-modal alignment. However, in the above process, two important issues are dismissed: temporal modeling and video-text partial misalignment.

Firstly, each frame is encoded independently as it passes through the visual encoder, which neglects the interactions between frames and hinders temporal understanding. To address this problem, existing research often introduces additional modules as either a posterior or intermediate structure for the visual encoder to explicitly incorporate temporal modeling for various downstream video tasks. For high-level semantic knowledge dominated tasks, \emph{i.e.,} video-language task, the posterior structure fully leverages the pretrained visual-language alignment knowledge by applying temporal modeling to the visual encoder output $\{ v_i \}_{i=1}^{T}$. Nevertheless, the highly semantic nature of ${ v_i }_{i=1}^{T}$ makes it challenging to capture low-level spatial-temporal patterns, leading to less effective temporal modeling. As for visual pattern dominated tasks, \emph{i.e.,} video-only task, the intermediate structure integrated within the visual encoder fully leverages the pretrained low-level visual patterns. This empowers the encoder with the capability of learning spatial-temporal patterns from the video. However, the plug-in modules disrupt the original model's structure and internal feature flow, resulting in the inability to inherit the high-level semantic information alignment capability from the pretrained models.

Secondly, the simple strategy in cross-modal interaction overlooks the prevalent issue of partial misalignment within video-text pairs. This misalignment results in aligned information being distributed selectively across specific frames and phrases, while other contextual elements may lack relevance to each other. The irrelevant parts are a kind of noise to video-language alignment. Therefore, simply representing the video and text with averaged representation or CLS embedding would introduce the noise hindering the learning of cross-modal alignment.

\begin{figure*}[h]
    \centering
    \setlength{\abovecaptionskip}{0.1 cm}
    \includegraphics[width=1\linewidth]{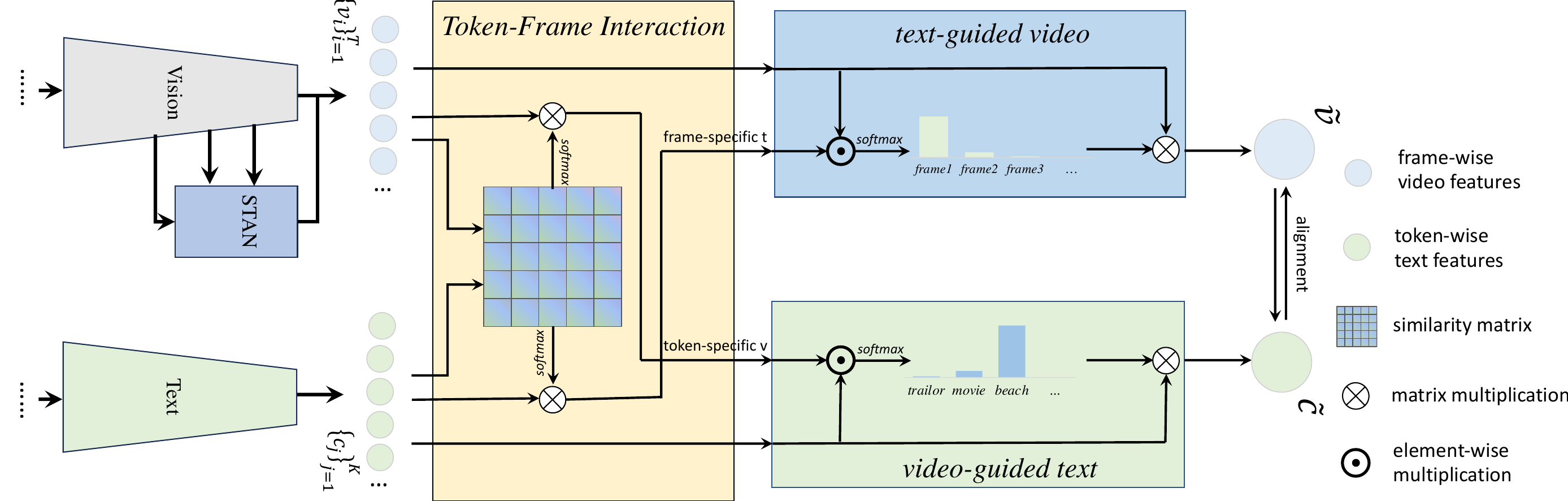}
    \caption{The overview of our proposed Mug. Based on the outputs of the video and text encoder, we first implement the mutual token-frame interaction on frame-wise video features and token-wise text features. Then, we compute the global video embedding and text embedding through guidance from another modality. Finally, we align the text-guided video embedding and video-guided text embedding.}
    \label{img_method2}
    \vspace{-1em}
\end{figure*}

In response to the issue of existing models not being able to simultaneously inherit the pretrained high-level and low-level knowledge, we introduce Spatial-Temporal Auxiliary Network (STAN), a novel temporal modeling mechanism for image-language pretrained models. As shown in Fig. \ref{img_method1}(middle), STAN functions as a branch structure alongside the pretrained visual encoder. With the sophisticated design, STAN leverages various levels of features while retaining the pretrained knowledge. The operation of STAN will be detailed in Sec. \ref{method2}. Additionally, as depicted in Figure \ref{img_method2}, to address the problem of partial misalignment, we introduce a novel cross-modal interaction module called Mutual-guided cross-modal alignment (Mug). This module takes frame-wise video representations $V$ and token-wise text representations $C$ as inputs. With guidance from the other modality, Mug efficiently filters out unrelated content and preserves aligned information in each modality, yielding new global video and text representation $\widetilde{v}$ and $\widetilde{c}$. Details about Mug will be provided in Section \ref{method3}.

\subsection{Spatial-Temporal Auxiliary Network} \label{method2}
Again, in the case of a video with $T$ frames, the frames are fed into the pretrained visual backbone, which generates intermediate outputs at the last $K+1$ levels of visual layers. We denote the outputs of the k$th$ selected visual layer as:
\begin{equation}
    V^k=\{f_{i,l}^{k} \in \mathcal{R}^{D}|i\in [1,T], l \in [0,L]\},
\end{equation}
which is a visual embedding sequence of the video where $T$, $L$ and $D$ represents the frame number, per-frame patch number and embedding dimension, respectively. In $V^k$, $f_{i,0}^{k}$ refers to the embedding of the [CLS] token in the $i$-th frame of the video, while $f_{i,l>0}^{k}$ represents the visual embedding of the $l$-th patch within that frame. Then, we take each intermediate output $V^k$ and pass it through the corresponding level of layer in STAN to model the spatial-temporal correspondence between video frames. At last, frame-wise outputs of the last pretrained visual layer are fuesed with the output of STAN to obtain the frame-wise video representation contextualized with temporal information, denoted as $\{ v_i \}_{i=1}^{T}$ in Eq. \ref{eq_represent}.

STAN is composed of a stack of $K$ spatial-temporal layers, with the input for each layer constructed upon the output of a pretrained vision layer and the last STAN layer. For the k$th$ layer in STAN, its input is an embedding sequence of the whole video denoted as:
\begin{equation}
    V'^k=\{f'^{k}_{0,0},f'^{k}_{1,1},..,f'^{k}_{1,L},..,f'^{k}_{T,1},..,f'^{k}_{T,L}\},
\end{equation}
where $f'^{k}_{0,0}$ is the embedding representing the whole video while others denote the embedding of image patches in different frames. The output of the STAN layer is also an embedding sequence maintaining the same size as its input, which is denoted as:
\begin{equation}
    \hat{V}^k = \{\hat{f}^{k}_{0,0},\hat{f}^{k}_{1,1},..,\hat{f}^{k}_{1,L},..,\hat{f}^{k}_{T,1},..,\hat{f}^{k}_{T,L}\}.
\end{equation}

At the first STAN layer, to construct its input from output of any pretrained visual layer $V^m$, we first average the embedding of [CLS] tokens in each frame as a new embedding $f'^{1}_{0,0}=\frac{1}{T}\sum_{i \in T}f^{m}_{i,0}$, and then update patch embeddings in $V^k$ with both spatial and temporal position embeddings as: 
\begin{equation} 
    f'^1_{i,l} = \mathrm{Dropout}(f_{i,l}^{m} + \mathrm{Pos_t}(t)+\mathrm{Pos_s}(l)),
\end{equation}
where $l>0$ and $\mathrm{Pos_t}$ and $\mathrm{Pos_s}$ are the learnable embeddings for the temporal and spatial positions of each patch. For the other layers in STAN, the input $V'^k$ is built based on the output from the previous STAN layer $\widetilde{V}^{k-1}$ and pretrained visual layer output $V^{m+k-1}$ as follows:
\begin{gather} 
    f'^{k}_{0,0}  = \widetilde{f}^{k-1}_{0,0} + \mathrm{W}^{k}_{proj} \frac{1}{T}\sum_{i \in T}f^{m+k-1}_{i,0}  , \\
    f'^{k}_{i,l}  = \widetilde{f}^{k-1}_{i,l} + \mathrm{W}^{k}_{proj} f^{m+k-1}_{i,l} , 
\end{gather}
where $i\in[1,T], l\in[1,L]$, and $\mathrm{W}^{k}_{proj} \in \mathbf{R}^{D \times D}$ is a projection layer. When compared to posterior structure based methods, STAN conducts spatial-temporal modeling on multi-level pretrained visual representations, enabling it to effectively capture visual dynamics information in the video. Meanwhile, unlike previous intermediate structure based methods that insert modules into pretrained visual encoder, STAN's branch structure protects the pretrained knowledge without disrupting the inherent encoder structure.

Given the input embedding sequence of a video, the STAN layer learns spatiotemporal information between video frames. As depicted in Fig. \ref{img_method1}(right), it performs temporal modeling through the alternating stacking of two independent modules -- the intra-frame module and the inter-frame module. Thanks to this separated design, we can reuse the structure of the pretrained visual encoder layer as our intra-frame spatial module and initialize it with the pre-trained parameter. This approach significantly reduces the optimization search space and improves the performance of downstream tasks. Same as most image-text pretrained models like CLIP, the intra-frame module is also a self-attention block designed for spatial modeling. To simplify notation, we omit the superscript of embedding and denote the embedding representation of the $i$-th frame as $X_i \in \mathbf{R}^{(L+1)\times D}$. Here, the embedding of the [CLS] token in the video is duplicated and concatenated with the patch embeddings. Within each frame, the spatial module updates the embeddings using self-attention:
\begin{equation}
    \hat{X}_i = \mathrm{softmax}(X_i\mathrm{W_Q} (X_i\mathrm{W_K})^\mathrm{T} / \sqrt{D})(X_i\mathrm{W_V}) + X_i,
\end{equation} 
where $\mathrm{W_Q} / \mathrm{W_K} / \mathrm{W_V}$ denote the linear projections for the query, key and value in self-attention layer of the spatial module. Afterward, the duplicated [CLS] embeddings in each frame are averaged to form the video [CLS] embedding. 

The cross-frame module is dedicated to temporal modeling. To simplify notation, we omit the superscript of the embedding and represent the collection of $l$-th patch embeddings in different frames as $Y_l \in \mathbf{R}^{T \times D}$. At each spatial position, the patch embeddings are updated using the function $Temp()$, which denotes the message passing strategy across temporal dimensions. In experiments, we will show that this strategy can be instantiated in various ways to facilitate temporal information exchange among frames. Here, we detail the instantiation of temporal self-attention, which possesses a natural advantage in sequence modeling. 
At each specific spatial position, the patch embeddings from different frames can be updated as: 
\begin{gather} 
    \label{qkv}
    \hat{Y}_l = \mathrm{W}_{proj} (\mathrm{softmax}(Y_l\mathrm{W_Q} (Y_l\mathrm{W_K})^\mathrm{T} / \sqrt{D})(Y_l\mathrm{W_V}) + Y_l), 
\end{gather}
where $\mathrm{W_Q} / \mathrm{W_K} / \mathrm{W_V}$ denote the linear projections for the query, key, and value in the self-attention layer of the cross-frame module, and $\mathrm{W}_{proj}$ is the extra temporal linear projection initialized as zero. By employing temporal attention, each patch in the video is contextualized with temporal information from the same locations, while the zero projection helps maintain training stability during the early stages.

At the final stage, with the output of the last pretrained visual layer $V^{-1}$ and the output of the last STAN layer $\hat{V}^K$, we can simply combine them through addition to form the ultimate output of the video encoder:
\begin{gather} 
    V = \mathrm{W}_{v\_proj}( \mathrm{LN}(V^{-1} \oplus \hat{V}^K)),
\end{gather}
where $\mathrm{LN}$ is the final layer normalization in pretrained visual encoder and $\mathrm{W}_{v\_proj}$ is the linear weight projecting the visual embedding into joint visual-text feature space. Furthermore, $\oplus$ means the global [CLS] token of STAN is duplicated T times and added to the [CLS] of each frame in $V^{-1}$, while the patch tokens are combined through simple addition. Finally, the same as the image encoder, we only have $L+1$ tokens for the video encoding. This property significantly reduces the computational burden if we need to further feed these tokens into multimodal encoders or LLMs, in comparison to the joint space-time video encoder \cite{xue2022clip, huang2023clover, bertasius2021space}. 

\subsection{Mutual-Guided Cross-Modal Alignment} \label{method3}
In the previous section, we have acquired the token-wise text embeddings $C$ and frame-wise video embeddings $V$. 
In this section, we will further delve into how to filter out misaligned information using Mug, as depicted in Fig. \ref{img_method2}. Mug first establishes token-frame-wise correspondences by calculating the dot-product similarity between $C$ and $V$. With the similarity matrix, we then introduce how to provide mutual guidance for feature aggregation from the perspective of each modality, respectively. 

From the perspective of video modality, we first filter out the most relevant information in the text for each video frame. This is achieved by calculating the \textit{frame-to-token attention distribution}, which assigns a score to each text token based on its relevance to the current video frame. Specifically, the attention score of the $i_{th}$ video frame with respect to the $j_{th}$ text token is given by:
 \begin{equation}
        s_{i,j} = \frac{\mathrm{exp}(\tau c_j \cdot v_i)}{\sum_{j=1}^K{\mathrm{exp}(\tau c_j \cdot v_i)}}, \label{eq3}
 \end{equation}
 where $\sum_{j=1}^K s_{i,j} = 1$, $\cdot$ represents dot-product operation and $\tau$ controls the sharpness of attention distribution. For example, in Fig. \ref{intro2}(a), the $1^{st}$ video frame is expected to have dominant attention on the tokens corresponding to the action of  ``\textit{people on a beach}'' across all the text tokens. 

Then, we aggregate the text embeddings based on the attention distribution and get the \textbf{\textit{frame-specific text embedding}} for each frame:
 \begin{equation}
        \overline{c_i} = \sum_{j=1}^K s_{i,j} c_j \label{eq4},~ \operatorname{where} ~ \overline{c_i} \in \mathbb{R}^{ D}.
 \end{equation}
 The set of \textit{frame-specific text embedding} $\{ \overline{c_i} \}_{i=1}^{T}$ represents updated text embeddings specified for each frame, where irrelevant information in the original text that is not aligned with the frame is suppressed and information that is relevant to the frame is strengthened.
 We use these updated text embeddings to evaluate the correspondence of each frame with respect to the text. This evaluation is done using dot-product similarity scores as the metric:
 \begin{equation}
        \widetilde{s_i} = \frac{\mathrm{exp}(\tau \overline{c_i} \cdot v_i)}{\sum_{n=1}^T{\mathrm{exp}(\tau \overline{c_n} \cdot v_n)}} \label{eq4},
 \end{equation}
 where $\widetilde{s_i}$ represents the attention weight of each frame towards the text. Through $\widetilde{s_i}$, we can further aggregate frame-wise embedding to the global video-level representation with the guidance of text. Formally, we define this global \textbf{\textit{text-guided video embedding}} as:
 \begin{equation}
        \widetilde{v} =\sum_{i=1}^T \widetilde{s_{i}} v_i, ~\operatorname{where} ~ \widetilde{v} \in \mathbb{R}^{ D}. \label{eq6}  
 \end{equation}

 Analogously, from the perspective of text modality, we follow the same procedure to get improved text embedding under the guidance of video. Specifically, we first calculate \textit{token-to-frame attention distribution}, which assigns a score to each frame embedding based on its relevance to the current text token:
 \begin{equation}
        s'_{i,j} = \frac{\mathrm{exp}(\tau c_j \cdot v_i)}{\sum_{i=1}^T{\mathrm{exp}(\tau c_j \cdot v_i)}}, \label{eq7}
 \end{equation}
 where $i$ and $j$ indicate the index of text token and frame.
Then, we get \textbf{\textit{token-specific video embedding}} for each text token to assess the token-to-video correspondence:
 \begin{equation}
        \overline{v_j} = \sum_{i=1}^T s'_{i,j} v_i,~ \operatorname{where}~ ~ \overline{v_i} \in \mathbb{R}^{ D}
 \end{equation}
 \begin{equation}
         \widetilde{s'_j} = \frac{\mathrm{exp}(\tau c_j \cdot \overline{v_j})}{\sum_{n=1}^K{\mathrm{exp}(\tau c_n \cdot \overline{v_n})}}, \label{eq8}
 \end{equation}
  where $\widetilde{s'_j}$ represents the attention weight of each text token towards the video. We obtain the global \textbf{\textit{video-guided text embedding}} $\widetilde{c}$ by aggregating the text token embeddings according to $\{\widetilde{s'_j}\}_{j=1}^K$:
\begin{equation}
    \widetilde{c} =\sum_{j=1}^K \widetilde{s'_{j}} c_j, ~\operatorname{where} ~ \widetilde{c} \in \mathbb{R}^{ D}, \label{eq10}
\end{equation}

In our proposed Mug, we default to using the frame-to-token interaction. However, our method can be readily adapted to various granularities of video-text interaction, such as video-to-token, frame-to-text, and token-to-token interaction. This flexibility allows for a trade-off between computation and interaction granularity, catering to different requirements based on the specific application. We will further explore and discuss this in our experiments.

\subsection{Training} 
\textit{Post-pretraining \& text-video retrieval.} Both post-pretraining and retrieval tasks utilize video-text pairs as training sources, resulting in the same training pipeline. Specifically, given \textit{text-guided video embedding} $\widetilde{v}$ and \textit{video-guided text embedding} $\widetilde{c}$, we calculate the dot-product similarity between the two embeddings, which serves as the similarity metric for the video and text in contrastive learning in a B-batch by:
\begin{equation}
\begin{split}
    & \mathcal{L}_{t2v} = - \frac{1}{B} \sum_{m=1}^B \mathrm{log} \frac{\mathrm{exp}(\tau \widetilde{c_{mn}} \cdot \widetilde{v_{nm}})}{\sum_{n=1}^B \mathrm{exp}(\tau \widetilde{c_{mn}} \cdot \widetilde{v_{nm}})} , \\
    & \mathcal{L}_{v2t} = - \frac{1}{B} \sum_{n=1}^B \mathrm{log} \frac{\mathrm{exp}(\tau \widetilde{v_{nm}} \cdot \widetilde{c_{mn}})}{\sum_{m=1}^B \mathrm{exp}(\tau \widetilde{v_{nm}} \cdot \widetilde{c_{mn}})},\\
    & \mathcal{L}_{co} = \mathcal{L}_{t2v} +     \mathcal{L}_{v2t} ,
\end{split}
 \end{equation}
 where $\widetilde{c_{mn}}$ and $\widetilde{v_{nm}}$ denotes the mutual-guided text/video embedding of the $m^{th}$ text and $n^{th}$ video in the batch, and $\mathcal{L}_{co}$ denotes the final contrastive loss. It is worth noting that the video and text embeddings have been normalized before computing Mug, thereby the normalization is not included in calculating the similarity.

 \textit{Video action recognition.} Different from video-language tasks, action recognition tasks have fixed textual labels. Hence, we freeze the text encoder and only train the video encoder during finetuning. Besides, we do not employ any additional prompt templates like ``a video of action \{ \}" \cite{radford2021learning, ni2022expanding} to wrap the tags. Then, we compute the loss with $\widetilde{v}$ and $\widetilde{c}$ as follows:
 \begin{equation}
     \mathcal{L}_{cr} = \sum_{n=1}^N y_n \mathrm{log} \frac{\mathrm{exp}(\tau \widetilde{v} \cdot \widetilde{c_n})}{\sum_{i=1}^N \mathrm{exp}(\tau \widetilde{v} \cdot \widetilde{c_i})},
 \end{equation}
 where $N$ is the class number, $y_n$ is the one-hot label for class $n$, $c_n$ is the value of class $n$ in global text embedding, and $\mathcal{L}_{cr}$ denotes the final cross-entropy loss.

 \textit{Video action detection.}
  Following the action detection pipeline in Slowfast \cite{Feichtenhofer_2019_ICCV} and VideoMAE \cite{tong2022videomae}, we add ROIAlign \cite{he2017mask} with MaxPooling to generate the regions of interest in the last layer, following a cross-entropy with sigmoid loss for multi-label prediction.

\begin{table*}[t]
\setlength{\tabcolsep}{3pt}
\centering
\caption{The zero-shot results of text-to-video retrieval and video recognition on six downstream datasets. Models exhibiting obvious unfair comparison are de-emphasized, \textit{i.e.,} involving extra modality, much larger models, or self-supervised pretraining.}
\resizebox{1.\textwidth}{!}{
\begin{tabular}{lcccc|cccc|cccc|ccc}
\toprule
\multirow{2}{*}{Method} & \multicolumn{4}{c|}{MSR-VTT} & \multicolumn{4}{c|}{DiDeMo} & \multicolumn{4}{c|}{LSMDC} & \multicolumn{1}{c}{HMDB-51} & \multicolumn{1}{c}{UCF-101} & \multicolumn{1}{c}{Kinetics400} \\

                                                & R@1   & R@5   & R@10 & MdR  & R@1   & R@5   & R@10 & MdR & R@1  & R@5  & R@10 & MdR  & Acc@1 & Acc@1 & Acc@1 \\
\hline
\emph{\textcolor{blue}{Non-CLIP models}}        &       &       &      &      &       &       &      &     &      &      &      &   &   &   &   \\
VideoCLIP~\cite{xu2021videoclip}                      & 10.4  & 22.2  & 30.0 & -    & 16.6  & 46.9  & -    & -   & -    & -    & -    & -  & - & - & -  \\
Frozen~\cite{bain2021frozen}                            & 18.7  & 39.5  & 51.6 & 10.0 & 21.1  & 46.0  & 56.2 & 7.0 & 9.3  & 22.0 & 30.1 & 51.0 & 27.5 & 45.4 & -\\
ALPRO~\cite{li2022align}                      & 24.1  & 44.7  & 55.4 & -    & 23.8  & 47.3  & 57.9 & -   & -    & -    & -    & -  & - & - & -  \\
VIOLET~\cite{fu2021violet}                    & 25.9  & 49.5  & 59.7 & -    & 23.5  & 49.8  & 59.8 & -   & -    & -    & -    & -  & - & - & -  \\
BridgeFormer~\cite{ge2022bridging}             & 26.0  & 46.4  & 56.4 & 7.0  & 25.6  & 50.6  & 61.1 & 5.0 & 12.2 & 25.9 & 32.2 & 42.0 & 38.0 & 51.1 & - \\
Clover~\cite{huang2023clover}             & 26.4  & 49.5  & 60.0 & 6.0  & 29.5  & 55.2  & 66.3 & 4.0 & 17.4 & 29.2 & 38.2 & 24.0 & - & - & - \\
OmniVL~\cite{wang2022omnivl} & 34.6  & 58.4  & 66.6 & -  & 33.4  & 58.7  & 68.5 & - & -    & -    & -    & - & - & - & - \\
\hline
\emph{\textcolor{blue}{CLIP-B/32}}             &       &       &      &      &       &       &      &     &      &      &      &    &  &  &   \\
CLIP~\cite{radford2021learning}                         & 30.6  & 54.4  & 64.3 & 4.0  & 24.7  & 49.3  & 60.9 & 6.0 & 13.6 & 27.9 & 35.5 & 32.0 & - & - & 42.1 \\
CLIP-straight~\cite{portillo2021straightforward}           & 31.2  & 53.7  & 64.2 & 4.0  & -     & -     & -    & -   & 11.3 & 22.7 & 29.2 & 56.5 & - & - & - \\
CLIP4Clip~\cite{luo2022clip4clip}                & 32.0  & 57.0  & 66.9 & 4.0  & -     & -     & -    & -   & 15.1 & 28.5 & 36.4 & 28.0 & - & - & - \\
BridgeFormer~\cite{ge2022bridging}             & 33.2  & 58.0  & 68.6 & 4.0  & -     & -     & -    & -   & 15.5 & 30.7 & 38.7 & 22.0 & - & - & - \\
CLIP-ViP~\cite{xue2022clip}                    & 29.0  & 51.2  & 61.3 & 5.0  & 22.6  & 43.9  & 56.4 & 7.0 & 11.3 & 25.3 & 31.3 & 38.0 & - & - & -\\
\rowcolor[RGB]{207,234,241} Mug-STAN-B/32               & \textbf{35.9}  & \textbf{60.8}  & \textbf{69.6} & \textbf{3.0}  & \textbf{33.7}  & \textbf{60.5}  & \textbf{70.3} & \textbf{3.0} & \textbf{17.4} & \textbf{32.7} & \textbf{40.4} & \textbf{21.5} & - & - & \textbf{48.1} \\
\hline
\emph{\textcolor{blue}{CLIP-B/16}}               &       &       &      &      &       &       &      &     &      &      &      &   &  &  &   \\
CLIP~\cite{radford2021learning}                           & 31.8  & 53.9  & 64.5 & 4.0  & 27.7  & 51.0  & 62.5 & 5.0 & 15.2 & 29.7 & 37.6 & 25.0 & 43.2 & 68.9 & 48.0 \\
ActionCLIP~\cite{wang2021actionclip}                    & -  & -  & - & -  & -  & -  & -  & -  & - & -  & -  & - & 40.8 & 58.3 & - \\
CLIP-ViP~\cite{xue2022clip}                    & 31.7  & 53.8  & 63.2 & 4.0  & 24.6  & 50.7  & 59.7 & 5.0 & 12.5 & 26.1 & 33.3 & 39.0 & 41.2 & 48.9 & 37.6 \\
X-CLIP~\cite{ni2022expanding}              & 31.7  & 53.8  & 63.2 & 4.0  & 24.6  & 50.7  & 59.7 & 5.0 & 12.5 & 26.1 & 33.3 & 39.0 & 44.6 & 72.0 & - \\
\rowcolor[RGB]{207,234,241}  Mug-STAN-B/16      & \textbf{38.7}  & \textbf{64.0}  & \textbf{74.0} & \textbf{2.0}  & \textbf{36.2}  & \textbf{62.3}  & \textbf{71.1} & \textbf{3.0} & \textbf{18.0} & \textbf{33.3} & \textbf{41.4} & \textbf{19.0} & \textbf{50.9} & \textbf{70.3} & \textbf{55.7} \\
\hline
\emph{\textcolor{blue}{CLIP-L/14}}           &       &       &      &      &       &      &      &     &      &      &      &   &  &  &   \\
CLIP~\cite{radford2021learning} &  35.4 & 58.8 & 68.1 & 3.0 & 30.3 & 54.9 & 65.4 & 4 & 18.5 & 33.8 & 42.3 & 19.0 & 46.5 & 72.7 & 55.9 \\
\textcolor{gray}{ImageBind$^*$~\cite{girdhar2023imagebind}} &  \textcolor{gray}{36.8} & \textcolor{gray}{61.8} & \textcolor{gray}{70.0} & \textcolor{gray}{-} & \textcolor{gray}{-} & \textcolor{gray}{-} & \textcolor{gray}{-} & \textcolor{gray}{-} & \textcolor{gray}{-} & \textcolor{gray}{-} & \textcolor{gray}{-} & \textcolor{gray}{-} & \textcolor{gray}{-} & \textcolor{gray}{-} & \textcolor{gray}{50.0} \\
\textcolor{gray}{InternVideo \cite{wang2022internvideo}} &  \textcolor{gray}{40.0} & \textcolor{gray}{65.3} & \textcolor{gray}{74.1} & \textcolor{gray}{2.0} & \textcolor{gray}{31.5} & \textcolor{gray}{57.6} & \textcolor{gray}{68.2} & \textcolor{gray}{3.0} & \textcolor{gray}{17.6} & \textcolor{gray}{32.4} & \textcolor{gray}{40.2} & \textcolor{gray}{23.0} & - & - & \textcolor{gray}{64.2} \\
\rowcolor[RGB]{207,234,241}  Mug-STAN-L/14                   & \textbf{41.7}  & \textbf{65.7}  & \textbf{75.8} & \textbf{2.0}  & \textbf{39.6}  & \textbf{64.3}  & \textbf{72.6} & \textbf{2.0} & \textbf{20.7} & \textbf{38.8} & \textbf{46.2} & \textbf{14.0} & \textbf{52.1} & \textbf{76.9} & \textbf{65.0} \\

\bottomrule
\end{tabular}}
\vspace{-1em}
\label{zero_shot}
\end{table*}

\section{Experiments}
\subsection{Datasets}
We evaluate our Mug-STAN on both video-language tasks, \textit{i.e.,}, video-text retrieval, and video-only tasks,  \textit{i.e.,}, video recognition and video detection, which trials our methods from the two different perspectives. For video-text retrieval, we use \textit{MSR-VTT} \cite{xu2016msr}, \textit{DiDemo} \cite{anne2017localizing} and \textit{LSMDC} \cite{rohrbach2017movie}; for video recognition, we use \textit{Kinetics-400} \cite{kay2017kinetics} and \textit{Something-Something-v2} \cite{goyal2017something}; for video detection, we adopt Atomic Visual Action V2.2 \cite{gu2018ava}. Besides, we conduct the video-text post-pretraining on datasets with different levels of noise, including WebVid10M \cite{bain2021frozen} and HowTo100M \cite{miech2019howto100m}. 

\textit{Video-Language Datasets}: \textit{MSR-VTT} is the most widely used benchmark for 
 video-text retrieval. It consists of 10,000 YouTube videos, each associated with 20 captions. We report our results on the 1K-A split \cite{yu2018joint}, which contains 9000 videos for training and 1000 for testing. \textit{DiDemo} includes 10,611 videos sourced from Flicker, accompanied by 40,000 sentences. Notably, this dataset features longer video durations compared to other retrieval datasets. Following previous works \cite{luo2022clip4clip, fang2021clip2video}, we concatenate all captions of a video into a single query. \textit{LSMDC} is a large-scale video-text retrieval benchmark comprising 118,081 videos sourced from 202 movies. This dataset offers a higher level of diversity in terms of concepts and video durations compared to other datasets.

\textit{Video-only Datasets}: \textit{Kinetics-400 (K-400)} is the most popular video recognition benchmark. Comprising over 300,000 video clips, Kinetics-400 covers 400 human action classes with average 300 frames. \textit{Something-Something-v2 (SSv2)} is a video action recognition benchmark specifically designed for temporal modeling capabilities. It consists of 220,485 videos, each associated with 174 action classes. In contrast, \textit{K-400} has a bias towards action categories with static scene context, as noted in \cite{Sevilla-Lara_2021_WACV}. However, in \textit{SSv2}, the action classes are less influenced by static scene context and instead focus more on dynamic information within the videos. \textit{Atomic Visual Action (AVA)} v2.2 is designed for spatial-temporal action detection. It provides dense annotation for 80 atomic visual actions across 430 15-minute movie clips, resulting in 1.62M action labels with multiple labels per human occurring frequently. 

\textit{Video Pretraining Datasets}: \textit{WebVid10M} is a large-scale video-text pretraining dataset of short videos with textual descriptions sourced from stock footage sites. With 
10.7M video-caption pairs and 52K total video hours, the videos are diverse and rich in their content, which has demonstrated fancy results in both downstream video-language tasks \cite{li2023videochat} and video generation tasks \cite{luo2023videofusion}. \textit{HowTo100M} is a large-scale dataset of narrated videos from Youtube videos. It features a total of 136M video clips with captions and 23k activities. Unlike Webvid, most of the captions in \textit{HowTo100M} are derived from automated speech recognition (ASR) or subtitles. Consequently, this leads to a more severe misalignment between the video and text.

\begin{table*}[t]
\setlength{\tabcolsep}{3pt}
\centering
\caption{The finetuning results of text-to-video retrieval on MSRVTT, DiDeMo, and LSMDC. Models exhibiting obvious unfair comparison are de-emphasized. For CLIP-based methods, * means extra tricks (\textit{e.g.,} DSL \cite{cheng2021improving} and QB-Norm \cite{bogolin2022cross}) are utilized during inference; and \dag \ denotes post-pretraining the models on video-text datasets before finetuning.}
\resizebox{1.\textwidth}{!}{
\begin{tabular}{lcccc|cccc|cccc}
\toprule
\multirow{2}{*}{Method} & \multicolumn{4}{c|}{MSR-VTT} & \multicolumn{4}{c|}{DiDeMo} & \multicolumn{4}{c}{LSMDC}  \\

                                                & R@1 $\uparrow$  & R@5 $\uparrow$  & R@10 $\uparrow$ & MdR $\downarrow$  & R@1 $\uparrow$  & R@5 $\uparrow$  & R@10 $\uparrow$ & MdR $\downarrow$ & R@1 $\uparrow$ & R@5 $\uparrow$ & R@10 $\uparrow$& MdR $\downarrow$ \\
\hline
\emph{\textcolor{blue}{Non-CLIP models}}        &       &       &      &      &       &       &      &     &      &      &      & \\
CLIPBert~\cite{lei2021less}                      & 22.0 & 46.8 & 59.9 & 6.0    & 20.4 & 48.0 & 60.8 & 6.0   & -    & -    & -    & -   \\
MMT~\cite{gabeur2020multi}                      & 26.6  & 57.1  & 69.6 & 4.0    & - & -  & - & -   & 12.9    & 29.9    & 40.1    & 19.3    \\
Frozen~\cite{bain2021frozen}                    &  31.0 & 59.5 & 70.5 & 3.0 &  31.0 & 59.8 & 72.4 & 3.0 &  15.0 & 30.8 & 40.3 & 20.0 \\
VIOLET~\cite{fu2021violet}                    & 34.5  & 63.0  & 73.7 & -    & 32.6  & 62.8  & 74.7 & -   & 16.1    & 36.6    & 41.2    & -   \\
HD-VILA~\cite{xue2022advancing}                 & 35.6 & 65.3 & 78.0 & 3.0  & 28.8 & 57.4 & 69.1 & 4.0   & 17.4 & 34.1 & 44.1 & 15.0  \\
All-in-one~\cite{wang2023all}                   &  37.9 & 68.1 & 77.1 & -   &  32.7 & 61.4 & 73.5 & 3.0  & -    & -    & -    & -   \\
BridgeFormer~\cite{ge2022bridging}             &  37.6 & 64.8 & 75.1 & 3.0  &  37.0 & 62.2 & 73.9 & 3.0  &  17.9 & 35.4 & 44.5 & 15.0  \\
Clover~\cite{huang2023clover}             &  40.5 & 68.8 & 79.4 & 2.0  &  50.1 & 76.7 & 85.6 & 1.0 & 24.8 & 44.0 & 54.5 & 8.0  \\
\hline
\emph{\textcolor{blue}{CLIP-B/32}}             &       &       &      &      &       &       &      &     &      &      &      &     \\
CLIP4Clip~\cite{luo2022clip4clip}            & 44.5 & 71.4 & 81.6 & 2.0  & 43.4 & 70.2 & 80.6 & 2.0   & 21.6 & 41.8 & 49.8 & 11.0  \\
CenterCLIP~\cite{zhao2022centerclip }        &  44.2 & 71.6 & 82.1 & 2.0  & -     & -     & -    & - &  21.7 & 39.8 & 49.8 & 11.0  \\
CAMoE*~\cite{cheng2021improving}           &  47.3 & 74.2 & 84.5 & 3.0   &  43.8 & 71.4 & - & -  &  25.9 & 46.1 & 53.7 & -  \\
CLIP2tv~\cite{gao2021clip2tv}             &  46.1 & 72.5 & 82.9 & 2.0 &  45.5 & 69.7 & 80.6 & 2.0 & 15.5 & 30.7 & 38.7 & 22.0 \\
ts2net~\cite{liu2022ts2}             &  47.0 & 74.5 & 83.8 & 2.0  &  41.8 & 71.6 & 82.0 & -   &  23.4 & 42.3 & 50.9 & 9.0 \\
DRL~\cite{wang2022disentangled}        &  47.4 & 74.6 & 83.8 & 2.0  &  47.9 & 73.8 & 82.7 & 2.0  &  24.9 & 45.7 & 55.3 & 7.0 \\
CLIP-ViP\dag~\cite{xue2022clip}         &  50.1 & 74.8 & 84.6 & 1.0  & 48.6 & 77.1 & 84.5 & 2.0 &  25.6 & 45.3 & 54.4 & 8.0 \\
CLIP-ViP*\dag~\cite{xue2022clip}       &  \textbf{55.9} & 77.0 & 86.8 & 1.0  &  53.8 & 79.6 & 86.5 & 1.0 &  26.0 & \textbf{46.4} & 54.9 & 8.0 \\
\rowcolor[RGB]{207,234,241} Mug-STAN-B/32    &  48.9 & 74.5 & 84.1 & 2.0 &  49.6 & 75.3 & 84.6 & 2.0  & 25.0 & 44.7 & 54.0 & 8.0 \\
\rowcolor[RGB]{207,234,241} Mug-STAN-B/32\dag   &  50.9 & 74.6 & 84.1 & 1.0  &  52.4 & 78.1 & 85.8 & 1.0 & 25.8  & 45.9  & 54.6 & 8.0  \\
\rowcolor[RGB]{207,234,241} Mug-STAN-B/32*\dag  & \textbf{55.9} & \textbf{77.3} & \textbf{88.0} & 1.0  &  \textbf{57.2} & \textbf{79.9} & \textbf{87.3} & 1.0  &  \textbf{26.9} & 46.0 & \textbf{55.4} & 7.0  \\
\hline
\emph{\textcolor{blue}{CLIP-B/16}}               &       &       &      &      &       &       &      &     &      &      &      &    \\
CenterCLIP~\cite{zhao2022centerclip }        &  48.4 & 73.8 & 82.0 & 2.0  & -     & -     & -    & -  &  24.2 & 46.2 & 55.9 & 8.0  \\
DRL*~\cite{wang2022disentangled}     &  53.3 & 80.3 & 87.6 & 1.0  &  49.0 & 76.5 & 84.5 & 2.0   &  26.5 & 47.6 & 56.8 & 7.0 \\
CLIP-ViP\dag~\cite{xue2022clip}      &  54.2 & 77.2 & 84.8 & 1.0  &  50.5 & 78.4 & 84.5 & 1.0 &  29.4 & 50.6 & 59.0 & 5.0 \\
CLIP-ViP*\dag~\cite{xue2022clip}     &  \textbf{57.7} & 80.5 & 88.2 & 1.0 &  55.3 & 82.0 & 89.3 & 1.0 &  30.7 & 51.4 & 60.6 & 5.0 \\
\rowcolor[RGB]{207,234,241} Mug-STAN-B/16    &  51.9 & 77.8 & 85.3 & 1.0  &  53.1 & 78.9 & 86.8 & 1.0  &  28.0 & 49.7 & 59.4 & 6.0 \\
\rowcolor[RGB]{207,234,241} Mug-STAN-B/16\dag   &  53.9 & 77.4 & 85.8 & 1.0  &  56.6 & 79.7 & 87.1 & 1.0 & 29.2 & 50.5 & 59.6 & 5.0  \\
\rowcolor[RGB]{207,234,241} Mug-STAN-B/16*\dag  &  57.3 & \textbf{81.6} & \textbf{88.4} & 1.0  & \textbf{61.2} & \textbf{84.1} & \textbf{88.9} & 1.0  &  \textbf{30.8} & \textbf{51.9} & \textbf{61.2} & 5.0  \\
\hline
\emph{\textcolor{blue}{CLIP-L/14}}           &       &       &      &      &       &      &      &     &      &      &      &    \\
\textcolor{gray}{InternVideo\dag \cite{wang2022internvideo}} &  \textcolor{gray}{55.2} & \textcolor{gray}{-} & \textcolor{gray}{-} & \textcolor{gray}{-} & \textcolor{gray}{57.9} & \textcolor{gray}{-} & \textcolor{gray}{-} & \textcolor{gray}{-} & \textcolor{gray}{34.0} & \textcolor{gray}{-} & \textcolor{gray}{-} & \textcolor{gray}{-}  \\
\rowcolor[RGB]{207,234,241} Mug-STAN-L/14\dag   &  56.6 & 80.0 & 87.8 & 1.0  &  60.4 & 85.0 & 90.5 & 1.0 & 35.3 & 58.0 & 65.2 & 3.0  \\
\rowcolor[RGB]{207,234,241} Mug-STAN-L/14*\dag  &  \textbf{61.3} & \textbf{82.6} & \textbf{90.1} & 1.0  & \textbf{65.0} & \textbf{87.5} & \textbf{92.0} & 1.0  &  \textbf{37.2} & \textbf{59.2} & \textbf{67.0} & 3.0  \\

\bottomrule
\end{tabular}}
\vspace{-1em}
\label{retrieval}
\end{table*}

\subsection{Experiment Settings}
\textit{Model Setting.} In most experiments, we adopt CLIP as the baseline image-language pretrained models for a fair comparison with previous works. For STAN, the number of STAN layers is set as 4 for all datasets except on SSv2 when it is set to 6. The STAN layers and CLIP layers are one-to-one corresponded from top to bottom. For Mug, we employ frame-to-token interaction by default. The temperature scalar $\tau$ in Mug is set to the same unlearnable value as the logit scale in CLIP because Mug does not change the scale of CLIP features during feature transformation.  To further evaluate the generalizability of Mug-STAN, we also implement Mug-STAN upon CoCa using the same configuration as CLIP.

\textit{Post-pretraining.} 
On both datasets, we employ a sparse sampling strategy \cite{lei2021less} to sample 12 frames with each frame resized to 224*224 for each video clip, and for text, the token length is set to 64. We use AdamW \cite{loshchilov2018fixing} optimizer with a weight decay of 0.001, and set the initial learning rate as 4e-6 and 4e-5 for \textit{CLIP} layers and STAN layers with a cosine annealing decay schedule. We train our model using only normalized contrastive loss and do not include other targets like masked language modeling or video-text matching. We train models with a batch size of 1024 for 3 epochs. It takes 1.6k GPU hours with 32 A100 GPUs for post-pretraining on HowTo100M, while the consumption is 0.8k GPU hours on WebVid10M. To evaluate the efficacy of post-pretraining, we compare the performance of post-pretrained models through both zero-shot and fine-tuning settings on downstream tasks.

\textit{Finetuning.} For all datasets, the batch size is set to 128, and we adopt AdamW as our optimizer with a weight decay of 0.02. For video-text retrieval, we adopt a frame number of 12 and a token length of 32 for MSRVTT, LSMDC. On Didemo where videos have a longer duration,  the frame number and token number are set to 64 and 64.  The learning rates are initialized to 2e-6 and 2e-5 for parameters in CLIP and STAN respectively. For video-only tasks, we sample 8 frames by default. The learning rates are initialized to 8e-6 and 8e-5 for CLIP and STAN layers.  For action detection, we further pretrain Mug-STAN on K400 following previous work, and adopt a frame span of 300, which aligns with the default frame number of Kinetics videos.

\begin{table*}[]
   \begin{center}
     \caption{The finetuning results of video recognition on Kinetics-400 and Something-Something-2. We present methods of comparable scale for fair comparison. We report the FLOPs of all views.} 
    \vspace{-1em}
     \footnotesize
     \label{recognition}
     \resizebox{1.\textwidth}{!}{
     \begin{tabular}{lccc|cc|cc} 
     \toprule
     \multicolumn{1}{c}{Methods}  & \multicolumn{1}{c}{Frames}  & \multicolumn{1}{c}{Testing Views} & \multicolumn{1}{c|}{GFLOPs} & \multicolumn{1}{c}{K400 Acc@1} & \multicolumn{1}{c|}{K400 Acc@5} & \multicolumn{1}{c}{SSv2 Acc@1} & \multicolumn{1}{c}{SSv2 Acc@5} \\ \hline 
     \emph{\textcolor{blue}{Non-CLIP models}}        &       &       &      &      &       &       &       \\  
     %\multicolumn{1}{l}{VTN-ViT-B \cite{neimark2021video}}  & 250 & $1 \times 1$ & 3992 & 78.6 & 93.7 & 78.6 & 93.7 \\
     \multicolumn{1}{l}{TimeSformer-L \cite{bertasius2021space}} & 96 & $1 \times 3$ & 7140 & 80.7 & 94.7 & 62.4 & - \\ 
     %\multicolumn{1}{l}{Mformer-HR \cite{patrick2021keeping}} & 16 & $10 \times 3$ & 28764 & 83.1 & 95.9 & 78.6 & 93.7 \\ 
     \multicolumn{1}{l}{Video-Swin-B \cite{liu2022video}} & 32 & $10 \times 5$ & 14729 & 82.7 & 95.5 & 69.6 & 92.7 \\ 
     \multicolumn{1}{l}{MViT \cite{fan2021multiscale}} &  32 & $3 \times 1$ & 1362 & 82.9 & 95.7 & 67.7 & 90.9 \\ 
     \multicolumn{1}{l}{ViViT-L \cite{arnab2021vivit}} &  32 & $4 \times 3$ & 11940 & 83.5 & 94.3 & 65.9 & 89.9 \\ 
     \multicolumn{1}{l}{MTV-B \cite{yan2022multiview}} &  32 & $4 \times 3$ & 11160 & 82.4 & 95.2 & 68.5 & 90.4 \\ 
     \hline 
     \emph{\textcolor{blue}{CLIP-B/16}}        &       &       &      &      &       &       &       \\  
     \multicolumn{1}{l}{\textit{CLIP}-B/16 \cite{radford2021learning}} & 8 & $4 \times 3$ & - & 81.1 & 94.8 & 44.0 & 76.8\\
     \multicolumn{1}{l}{Action-CLIP-B/16 \cite{wang2021actionclip}} & 32 & $10 \times 3$ & 16890 & 83.8 & 96.2 & - & - \\  
     \multicolumn{1}{l}{A6 \cite{ju2022prompting}} & 16 & $ - $ & - & 76.9 & 93.5 & - & - \\  
     \multicolumn{1}{l}{STadapter-CLIP-B/16 \cite{pan2022st}} & 8 & $1 \times 3$ & 455 & 82.0 & 95.7 & 67.1 & 91.2 \\
     \multicolumn{1}{l}{STadapter-CLIP-B/16 \cite{pan2022st}} & 32 & $1 \times 3$ & 1821 & 82.7 & 96.2 & \textbf{69.5} & 92.6 \\ 
     \multicolumn{1}{l}{X-CLIP-B/16 \cite{ni2022expanding}} & 8 & $4 \times 3$ & 1740 & 83.8 & 96.7 & 63.1 & 89.0 \\
     \multicolumn{1}{l}{X-CLIP-B/16 \cite{ni2022expanding}} & 16 & $4 \times 3$ & 3444 & 84.7 & 96.8 & - & - \\ 
     \rowcolor[RGB]{207,234,241} \multicolumn{1}{l}{Mug-STAN-B/16} & 8 & $1 \times 3$ & 593  & 84.7 & 96.7 & 67.7 & 91.5 \\
     \rowcolor[RGB]{207,234,241} \multicolumn{1}{l}{Mug-STAN-B/16} & 16 & $1 \times 3$ & 1187  & \textbf{85.1} & \textbf{96.9} & \textbf{69.5} & \textbf{92.8} \\ 
     \bottomrule
     \vspace{-1.8em}
   \end{tabular}  }
   \end{center}
\end{table*}

\begin{table}[]
    \setlength{\abovecaptionskip}{0.cm}
    \setlength{\belowcaptionskip}{-0.3cm}
    \begin{center}
      \caption{The finetuning results of video detection on AVA 2.2. Models utilizing self-supervised reconstruction are de-emphasized. * means our implementation.} 
      
      \label{ava}
      \renewcommand\tabcolsep{9pt}
      \resizebox{1.\columnwidth}{!}{
      \begin{tabular}{l|c|c|c|c} 
      \toprule
      \multicolumn{1}{l|}{Method} & \multicolumn{1}{c|}{Pretrain} & \multicolumn{1}{c|}{Frames} & \multicolumn{1}{c|}{GFLOPs} & \multicolumn{1}{c}{mAP} \\ \hline 
      \multicolumn{1}{l|}{SlowFast \cite{Feichtenhofer_2019_ICCV}}  & K400 & \multicolumn{1}{c|}{32} & 138 &  23.8 \\
      \multicolumn{1}{l|}{MViTv1-B \cite{fan2021multiscale}}  & K400 & \multicolumn{1}{c|}{64} & 455 & 27.3 \\
      \multicolumn{1}{l|}{MViTv2-B \cite{Li_2022_CVPR}}  & K400 & 32 & 255 & 28.1 \\
      \textcolor{gray}{MVD-B \cite{wang2023masked}} & \textcolor{gray}{K400} & \textcolor{gray}{8} & \textcolor{gray}{180} & \textcolor{gray}{29.8} \\
      \textcolor{gray}{VideoMAE-B \cite{tong2022videomae}} & \textcolor{gray}{K400} & \textcolor{gray}{8} & \textcolor{gray}{180} & \textcolor{gray}{31.8} \\   
      \multicolumn{1}{l|}{CLIP-B/16* \cite{radford2021learning}}  & K400 & 8 & 180 & 24.9 \\
      \multicolumn{1}{l|}{XCLIP-B/16* \cite{ni2022expanding}}  & K400 & 8 & 185 & 27.6 \\
      \rowcolor[RGB]{207,234,241} \multicolumn{1}{l|}{Mug-STAN-B/16 }  & K400 & \multicolumn{1}{c|}{8} & 197 & \textbf{29.3} \\  
      \bottomrule
    \end{tabular}  }
    \end{center}
    \vspace{-1.5em}
  \end{table}

\vspace{-0.5em}  

\subsection{Comparison With State-of-the-Art Methods}
\textit{Zero-Shot Results.} 
The zero-shot results of WebVid10M post-pretraining are posted in Table. \ref{zero_shot}. We evaluate Mug-STAN on three text-video retrieval datasets and three video recognition datasets. We report our results under different model capacities, including on \textit{CLIP-B/32}, \textit{CLIP-B/16}, and \textit{CLIP-L/14}. As evident from the presentation, numerous approaches that introduce new structures onto CLIP tend to compromise its zero-shot capabilities, despite achieving improved fine-tuning outcomes, such as ActionCLIP, CLIP-ViP, and XCLIP. In contrast, Mug-STAN demonstrates clear zero-shot advantages over CLIP following post-pretraining. Note that our comparison with CLIP is conducted fairly, considering the little improvement achieved through CLIP post-pretraining detailed in Table \ref{zero_shot}. Moreover, in comparison to the previous SOTA methods in the zero-shot setting, our approach demonstrates significant advantages across all datasets, even when the comparisons are conducted unfairly for us. For instance, InternVideo \cite{wang2022internvideo} utilizes dual visual encoders, and generative self-supervised techniques, and involves 50 times more GPU days compared to our approach. Nevertheless, our method outperforms InternVideo by significant margins, achieving improvements of 1.7\%, 8.1\%, 3.1\%, and 0.8\% on the MSRVTT, DiDeMo, LSMDC, and Kinetics400 datasets, respectively. The results demonstrate our post-pretraining on Mug-STAN does not damage the rich knowledge in the CLIP while providing a stronger zero-shot capacity for video tasks. 

\textit{Video-Language Tasks.} 
We report the finetuning results of text-to-video retrieval in Table \ref{retrieval}. We compare our Mug-STAN with current SOTAs with various setting, including directly finetuning, finetuning after post-pretraining and using extra tricks during inference. As demonstrated in the results, when directly fine-tuning for video-text retrieval tasks, Mug-STAN brings about obvious advantage over CLIP, outperforming CLIP4clip by 4.7\% at R@1 on average across the three datasets with \textit{CLIP}-B/32 as backbone. Compared to another state-of-the-art method \textit{DRL} \cite{wang2022disentangled}, which also leverages frame-token wise interaction to boost performance, Mug-STAN outperforms it by 1.1\% at R@1 on average across the three datasets. When it comes to post-pretraining,  it is worth noting that only a few methods \cite{xue2022clip,luo2022clip4clip,wang2022internvideo} have explored this area, with \textit{CLIP-ViP} \cite{xue2022clip} being the strongest competitor. Compared to \textit{CLIP-ViP }, which introduces an external strong captioner \cite{wang2022ofa} to augment pre-training datasets with additional captions, our method is free from such complex data augmentation and achieves competitive or even better performance across different datasets. Moreover, Mug-STAN is able to bring about performance gains by post-pretraining on smaller or noisier datasets, while \textit{CLIP-ViP} requires larger dataset \textit{i.e.,} HDVilla-100M \cite{xue2022advancing}. Furthermore, compared to large competitors \cite{wang2022internvideo}, despite the disadvantages in terms of training cost, pretraining method, and model scale, MugSTAN still outperforms InterVideo across the three datasets.

\vspace{0.2em}

\begin{table*}[]
    \setlength{\abovecaptionskip}{0.cm}
    \begin{center}
      \caption{Ablation results of different components in our model on different settings. ``FT'' means direct finetuning results without pertaining; ``ZS'' means the zero-shot result after pertaining. We report the result of Recall@1.} 
      \vspace{-0.5em}
      \label{ablation_model}
      \renewcommand\tabcolsep{9pt}
      \resizebox{1.\textwidth}{!}{
      \begin{tabular}{cccccccc} 
      \toprule
      \multicolumn{4}{c|}{Components} & \multicolumn{4}{c}{Results} \\ 
      \multicolumn{1}{c}{Branch structure} & \multicolumn{1}{c}{Multi-level} & \multicolumn{1}{c}{Separated-ST}  &  \multicolumn{1}{c|}{Mutual-Guided} & \multicolumn{1}{c}{FT-MSRVTT} & \multicolumn{1}{c}{FT-DiDemo} & \multicolumn{1}{c}{ZS-MSRVTT} & \multicolumn{1}{c}{ZS-DiDemo}\\ \hline \hline 
       & & & \multicolumn{1}{c|}{}  & 43.1  & 43.4 & 30.6 & 24.7\\
       $\checkmark$ & $\checkmark$ & $\checkmark$ & \multicolumn{1}{c|}{} & 46.9 & 46.2 & 33.0 & 28.1\\ 
        & &  & \multicolumn{1}{c|}{$\checkmark$} & 46.1 & 45.4 & 33.1 & 29.8 \\ 
        & & $\checkmark$ & \multicolumn{1}{c|}{} & 44.9 & 43.5 & 31.9 & 25.4 \\ 
      $\checkmark$ &  & $\checkmark$ & \multicolumn{1}{c|}{} & 44.2 & 43.6 & 31.8 & 25.4 \\ 
      $\checkmark$ & $\checkmark$ &  & \multicolumn{1}{c|}{}  & 45.5 & 44.7 & 32.2 & 26.9\\ 
      $\checkmark$ & $\checkmark$ & $\checkmark$ & \multicolumn{1}{c|}{$\checkmark$}  & \textbf{48.9}  & \textbf{49.6}  & \textbf{35.9} & \textbf{33.7} \\ \bottomrule
    \end{tabular}  }
    \end{center}
    \vspace{-1em}
  \end{table*}

  \begin{table*}[]
   \setlength{\abovecaptionskip}{0.cm}
   \setlength{\belowcaptionskip}{-0.3cm}
   \begin{center}
     \caption{Ablation results on the post-pretraining. We report the finetuning results after post-pretraining with CLIP-B/32 on both MSR-VTT and DiDemo. We conduct the pretraining with different methods and pretraining datasets.}  
     \vspace{-0.5em}
     \footnotesize
     \label{ablation_training}
     \renewcommand\tabcolsep{9pt}
     \resizebox{1.\textwidth}{!}{
     \begin{tabular}{cccccccccc} 
     \toprule
     \multicolumn{1}{c}{\multirow{2}{*}{Model}} & \multicolumn{1}{c|}{\multirow{2}{*}{Pretrain Dataset}}  & \multicolumn{4}{c|}{DiDemo}  & \multicolumn{4}{c}{MSR-VTT} \\ 
      & \multicolumn{1}{c|}{} &  R@1 & R@5 & R@10 & \multicolumn{1}{c|}{MdR} &  R@1 & R@5 & R@10 & MdR \\ \hline 
      \textit{CLIP} & \multicolumn{1}{c|}{-} & 43.4 & 70.9 & 79.2 & \multicolumn{1}{c|}{2} & 43.1 & 69.8 & 81.8 & 2 \\
      \textit{CLIP} &  \multicolumn{1}{c|}{HowTo100M} & 43.0 & 70.2 & 80.3 & \multicolumn{1}{c|}{2} & 43.4 & 69.7 & 82.3 & 2 \\
      \textit{CLIP} &  \multicolumn{1}{c|}{WebVid10M} & 43.6 & 70.4 & 80.5 & \multicolumn{1}{c|}{2} & 43.9 & 70.1 & 80.0 & 2 \\ \hline 
      STAN+CLIP & \multicolumn{1}{c|}{-} & 46.5 & 71.5 & 80.9 & \multicolumn{1}{c|}{2} & 46.9 & 72.8 & 82.8 & 2 \\
      STAN+CLIP &  \multicolumn{1}{c|}{HowTo100M} & 47.0 & 72.1 & 81.6 & \multicolumn{1}{c|}{2} & 47.1 & 72.3 & 82.5  & 2 \\
      STAN+CLIP &  \multicolumn{1}{c|}{WebVid10M}  & 48.2 & 76.7 & 85.0 & \multicolumn{1}{c|}{2} & 47.5 & 72.8 & 82.9 & 2 \\ \hline 
      Mug+STAN+CLIP & \multicolumn{1}{c|}{-} & 49.6 & 75.3 & 84.6 & \multicolumn{1}{c|}{2} & 48.9 & 74.5 & 84.1 & 2 \\
      Mug+STAN+CLIP &  \multicolumn{1}{c|}{HowTo100M} & 51.6 & 77.3 & 85.2 & \multicolumn{1}{c|}{1} & 50.0 & 75.1 & 83.6 & 2 \\
      Mug+STAN+CLIP &  \multicolumn{1}{c|}{WebVid10M}  & 52.4 & 78.1 & 85.8 & \multicolumn{1}{c|}{1} & 50.9 & 74.6 & 84.1 & 1\\ 
      \bottomrule
     
   \end{tabular}  }
   \end{center}
   \vspace{-1.5em}
\end{table*} 

\begin{table}[]
\renewcommand\arraystretch{1.2}
   \setlength{\abovecaptionskip}{0.cm}
   \begin{center}
     \caption{Finetuning results of Mug-STAN on CoCa \cite{yu2022coca} on MSR-VTT and DiDemo retrieval. \dag \ denotes finetuning after post-pretraining.} 
     \vspace{-0.5em}
     \footnotesize
     \label{coca}
     \renewcommand\tabcolsep{9pt}
     \resizebox{1.\columnwidth}{!}{
     \begin{tabular}{ccccc} 
      \toprule
      \multicolumn{1}{c|}{Model}  & \multicolumn{1}{c}{R@1} & \multicolumn{1}{c}{R@5} & \multicolumn{1}{c}{R@10} & \multicolumn{1}{c}{MdR}\\ \hline 
      & \multicolumn{4}{c}{MSR-VTT} \\ \hline 
      \multicolumn{1}{l|}{CoCa (Baseline)} & 42.7 & \multicolumn{1}{c}{70.2} & 79.2 & 2.0 \\
      \multicolumn{1}{l|}{STAN-CoCa}  & 44.7 & \multicolumn{1}{c}{72.9} & 81.9 &  2.0 \\
      \multicolumn{1}{l|}{Mug-STAN-CoCa}  & 46.2 & \multicolumn{1}{c}{73.4} & 82.3 & 2.0 \\
      \multicolumn{1}{l|}{Mug-STAN-CoCa \ \dag}  & \textbf{48.0} & \multicolumn{1}{c}{\textbf{73.9}} & \textbf{82.4} & 2.0 \\ \hline 
       & \multicolumn{4}{c}{DiDemo} \\ \hline 
      \multicolumn{1}{l|}{CoCa (Baseline)} & 39.6 & \multicolumn{1}{c}{67.9} & 77.0 & 2.0 \\
      \multicolumn{1}{l|}{STAN-CoCa}  & 43.5 & \multicolumn{1}{c}{72.9} & 82.0 &  2.0 \\
      \multicolumn{1}{l|}{Mug-STAN-CoCa}  & 46.7 & \multicolumn{1}{c}{73.9} & 82.0 & 2.0 \\
      \multicolumn{1}{l|}{Mug-STAN-CoCa  \ \dag}  & \textbf{48.8} & \multicolumn{1}{c}{\textbf{73.8}} & \textbf{83.7} & 1.5 \\ \bottomrule
    \end{tabular} }
   \end{center}
   \vspace{-0.5em}
\end{table}  

\textit{Video-Only Tasks.} 
We report the finetuning results of video recognition and video detection on Kinetics-400, Something-Something-2, and AVA-v2.2 in Table \ref{recognition} and \ref{ava} respectively.
In the K400-recognition benchmark, CLIP-based methods demonstrate competitive performance with smaller model scales compared to image-pretrained methods. For instance, our VIT-B/16 based STAN achieves superior results compared to models like ViViT \cite{arnab2021vivit} and Video-swin \cite{liu2022video}, which have more than 10× GFLOPs compared to our method.
As for SSv2 and AVA benchmark, we observe that, without temporal modeling, bare CLIP model \cite{radford2021learning} achieves only 44.0\% top-1 accuracy and 25.9 mAP which dramatically under-performs ImageNet-Kinetics pretrained models, though it owns pretrained knowledge obtained from a much larger image-text dataset. The result suggests that the domain gap is significant between SSv2/AVA and CLIP model, and temporal modeling capability is desired for the two datasets. STAN brings about more than 25.5\% and 4.4\% performance improvement over the CLIP baseline on SSv2 and AVA, which demonstrates that Mug-STAN empowers CLIP with strong temporal modeling capability. It is worth noting that, in comparison to video-language tasks, the contrastive video-text pretraining does not demonstrate significant advantages over image-pretraining on video-only tasks. This is particularly evident for self-supervised reconstruction methods. Nevertheless, Mug-STAN manages to achieve competitive performance even in the face of this challenge when compared to single-modality pretrained methods. Moreover, in comparison to other CLIP-based methods, Mug-STAN consistently exhibits advantages across various datasets.

\begin{figure*}[h]
    \centering
    \setlength{\abovecaptionskip}{0.1 cm}
    \includegraphics[width=1\linewidth]{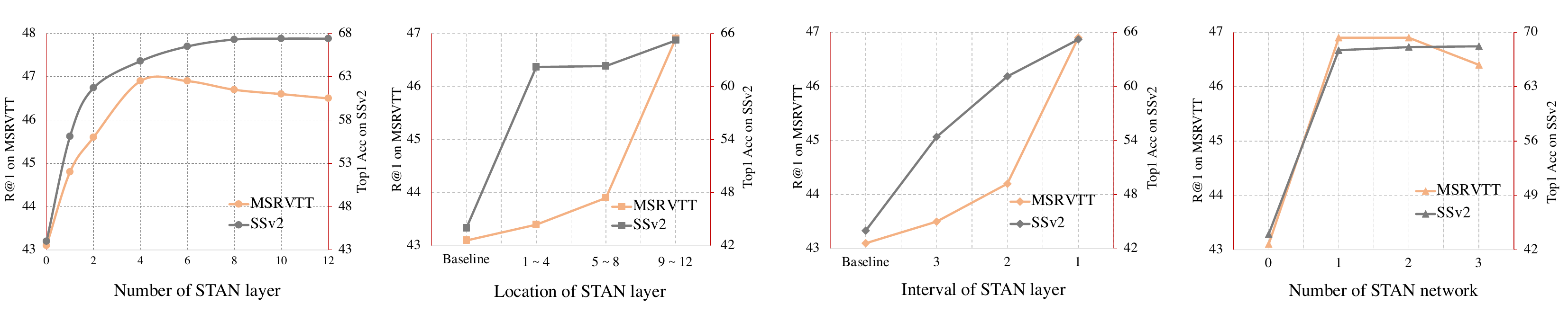}
    \vspace{-1.5em}
    \caption{Ablation results on the hyper-parameter setting of STAN. We report the finetuning results without post-pretraining on both MSRVTT and SSv2. We study the number of STAN layers, the relative location of STAN layer respect to CLIP, the interval of STAN layer (\textit{i.e.}, the number of CLIP layers between STAN layer), and the number of STAN networks.}
    \label{stan_ablation}
    \vspace{-1em}
\end{figure*}

\begin{table}[]
    \setlength{\abovecaptionskip}{0.cm}
    \setlength{\belowcaptionskip}{-0.3cm}
    \begin{center}
      \caption{Ablation results on the interaction module, including interaction granularity (middle) and interaction strategies (bottom). We report the results on both MSRVTT and DiDemo.} 
      \vspace{-1em}
      \label{ablation_mug}
      \renewcommand\tabcolsep{9pt}
      \resizebox{1.\columnwidth}{!}{
      \begin{tabular}{ccccc} 
      \toprule
      \multicolumn{1}{c|}{\multirow{2}{*}{Model}} & \multicolumn{2}{c|}{MSR-VTT} & \multicolumn{2}{c}{DiDemo} \\ 
      \multicolumn{1}{c|}{} & \multicolumn{1}{c}{R@1} & \multicolumn{1}{c|}{mean} & \multicolumn{1}{c}{R@1} & \multicolumn{1}{c}{mean}\\ \hline 
      \multicolumn{1}{l|}{STAN (Baseline)}  & 46.9 & \multicolumn{1}{c|}{67.5} & 46.5 &  66.3 \\ \hline
      \multicolumn{1}{l|}{Frame-Text Interaction}  & 47.5 & \multicolumn{1}{c|}{68.2} & 48.7 & 68.3 \\
      \multicolumn{1}{l|}{Video-Token Interaction}  & 46.8 & \multicolumn{1}{c|}{68.0} & 47.2 & 67.1 \\
      \multicolumn{1}{l|}{Frame-Token Interaction}  & \textbf{48.9} & \multicolumn{1}{c|}{\textbf{69.2}} & \textbf{49.6} & \textbf{69.8} \\ \hline
      \multicolumn{1}{l|}{Mug-STAN (max)}  & 47.3 & \multicolumn{1}{c|}{68.5} & 47.9 & 68.2 \\
      \multicolumn{1}{l|}{Mug-STAN (top3)}  & 48.6 & \multicolumn{1}{c|}{69.1} & 48.5 & 68.7 \\ 
      \multicolumn{1}{l|}{Mug-STAN (top5)}  & 48.1 & \multicolumn{1}{c|}{68.2} & 49.4 & 69.3 \\ 
      \multicolumn{1}{l|}{Mug-STAN (top7)}  & 47.0 & \multicolumn{1}{c|}{67.8} & 48.4 & 68.3 \\ 
      \multicolumn{1}{l|}{WTI-STAN \cite{wang2022disentangled}}  & 47.5 & \multicolumn{1}{c|}{68.8} & 47.2 & 67.9 \\ 
      \multicolumn{1}{l|}{Hunyuan-STAN \cite{jiang2022tencent}}  & 47.5 & \multicolumn{1}{c|}{68.8} & 47.5 & 67.4 \\  
      \multicolumn{1}{l|}{Mug-STAN (softmax)}  & \textbf{48.9} & \multicolumn{1}{c|}{\textbf{69.2}} & \textbf{49.6} & \textbf{69.8} \\  \bottomrule  
    \end{tabular}  }
    \end{center}
    \vspace{-1em}
  \end{table}
  
\subsection{Ablation Study}
\textit{Ablations on components of Mug-STAN.} 
To evaluate the contribution of different components in our method, we conduct ablation experiments on both finetuning setting and zero-shot setting as shown in Table. \ref{ablation_model}. First of all, in the first three lines are the overall performance of STAN and Mug, we can conclude that STAN and Mug are compatible with each other while each of them contributes to the adaption of image-language pretraining models, \textit{i.e.}, Mug addresses the issue of  partial misalignment in video-text data and STAN focuses on the temporal modeling. Moreover, combining Mug and STAN, the performance is further increased by a considerable margin, which demonstrates that the temporal modeling capability and the addressing of partial misalignment are mutually beneficial to each other. Secondly, lines 4-7 demonstrate the internal structure of STAN. Specifically, when we eliminate the branch structure or multi-level feature learning, the performance of STAN experiences a substantial decline across all four benchmarks. This serves as strong evidence of the superiority of our model structure over the posterior structure. 
Additionally, adopting joint-ST temporal modeling in STAN also brings noticeable improvements, albeit not surpassing the separate approach, which underscores the significance of reusing parameters from the pretrained model.

\vspace{0.2em}
  
\textit{Ablations on Post-Pretraining.} 
CLIP-ViP \cite{xue2022clip} points out two factors that potentially hinder the video post-pretraining to further improve the performance on downstream video tasks: \textit{dataset scale} and \textit{domain gap}. In this paper, through ablation study on post-pretraining, we figure out that empowering the pretrained model with \textit{temporal modeling capability}  and \textit{addressing partial-misalignment problem} are also crucial for post-pretraining. We employ HowTo100M and WebVid10M as pretraining dataset and train different models on the two datasets, respectively. 

\begin{table}[]
    \setlength{\abovecaptionskip}{0.cm}
    \setlength{\belowcaptionskip}{-0.3cm}
    \begin{center}
      \caption{Ablation results on the effectiveness of Mug in mitigating partial misalignment. Above, we report the R@1 scores for data with varying degrees of misalignment in MSRVTT and DiDemo datasets. Below, we compare Mug-STAN with other state-of-the-art video denoising methods.} 
      \vspace{-0.2em}
      \renewcommand\tabcolsep{9pt}
      \resizebox{1.\columnwidth}{!}{
      \begin{tabular}{ccccccc} 
      \toprule
      \multicolumn{1}{c|}{\multirow{2}{*}{level}} & \multicolumn{3}{c|}{MSR-VTT} & \multicolumn{3}{c}{DiDemo} \\ 
      \multicolumn{1}{c|}{} & \multicolumn{1}{c}{\%} & \multicolumn{1}{c}{STAN} & \multicolumn{1}{c|}{+Mug} & \multicolumn{1}{c}{\%} & \multicolumn{1}{c}{STAN}  & \multicolumn{1}{c}{+Mug}\\ \hline 
      \multicolumn{1}{l|}{top}  & 38 & 53.7 & \multicolumn{1}{c|}{54.0} & 42 & 54.3 &  54.8 \\
      \multicolumn{1}{l|}{middle}  & 23 & 47.2& \multicolumn{1}{c|}{49.0} & 23 & 43.3 &  46.9 \\
      \multicolumn{1}{l|}{bottom}  & 39 & 40.1 & \multicolumn{1}{c|}{43.0} & 35 & 39.2 &  45.1 \\
      \multicolumn{1}{l|}{all}  & 100 & 46.9 & \multicolumn{1}{c|}{48.9} & 100 & 46.5 &  49.6 \\ \hline \hline
      \multicolumn{1}{c|}{Model} & \multicolumn{3}{c|}{R@1 on HTM-Align} & \multicolumn{3}{c}{R@1 on YouCook2} \\  \hline 
      \multicolumn{1}{l|}{MIL-NCE \cite{miech2020end}}  & \multicolumn{3}{c|}{31.3} & \multicolumn{3}{c|}{15.1} \\
      \multicolumn{1}{l|}{TAN \cite{han2022temporal}}  & \multicolumn{3}{c|}{49.4} & \multicolumn{3}{c|}{20.1} \\
      \multicolumn{1}{l|}{Mug-STAN}  & \multicolumn{3}{c|}{\textbf{51.6}} & \multicolumn{3}{c|}{\textbf{29.7}}   \\
      \bottomrule  
    \end{tabular}  }
    \end{center}
    \label{ablation_misalign}
    \vspace{-1em}
  \end{table}
  
As shown in Table \ref{ablation_training}, for the \textit{CLIP} baseline, which employs a simple mean pooling strategy for cross-frame modeling, it takes trivial advantages from post-pretraining. As for experiments on STAN, which owns expertise on temporal modeling, we observe that post-pretraining on WebVid10M brings more performance gains than that on \textit{CLIP} baseline. When it comes to \textit{Mug-STAN}, the performance gains of post-pretraining on WebVid10M further increase to 2.8\% on DiDemo and 2.0\% on MSR-VTT. Moreover, even on HowTo100M, which consists of instructional videos with noise narrations and suffers from extremely severe partial-misalignment problem, our method still brings about 2.0\% and 1.1\% performance gains on DiDemo and  MSR-VTT, respectively. The results reveal that temporal modeling capability is beneficial to the post-pretraining while addressing partial-misalignment problem is able to further amplify the performance gains remarkably.

\vspace{0.2em}

\begin{figure}[h] 
    \setlength{\abovecaptionskip}{-0.1 cm}
    \centering
    \includegraphics[width=1\columnwidth]{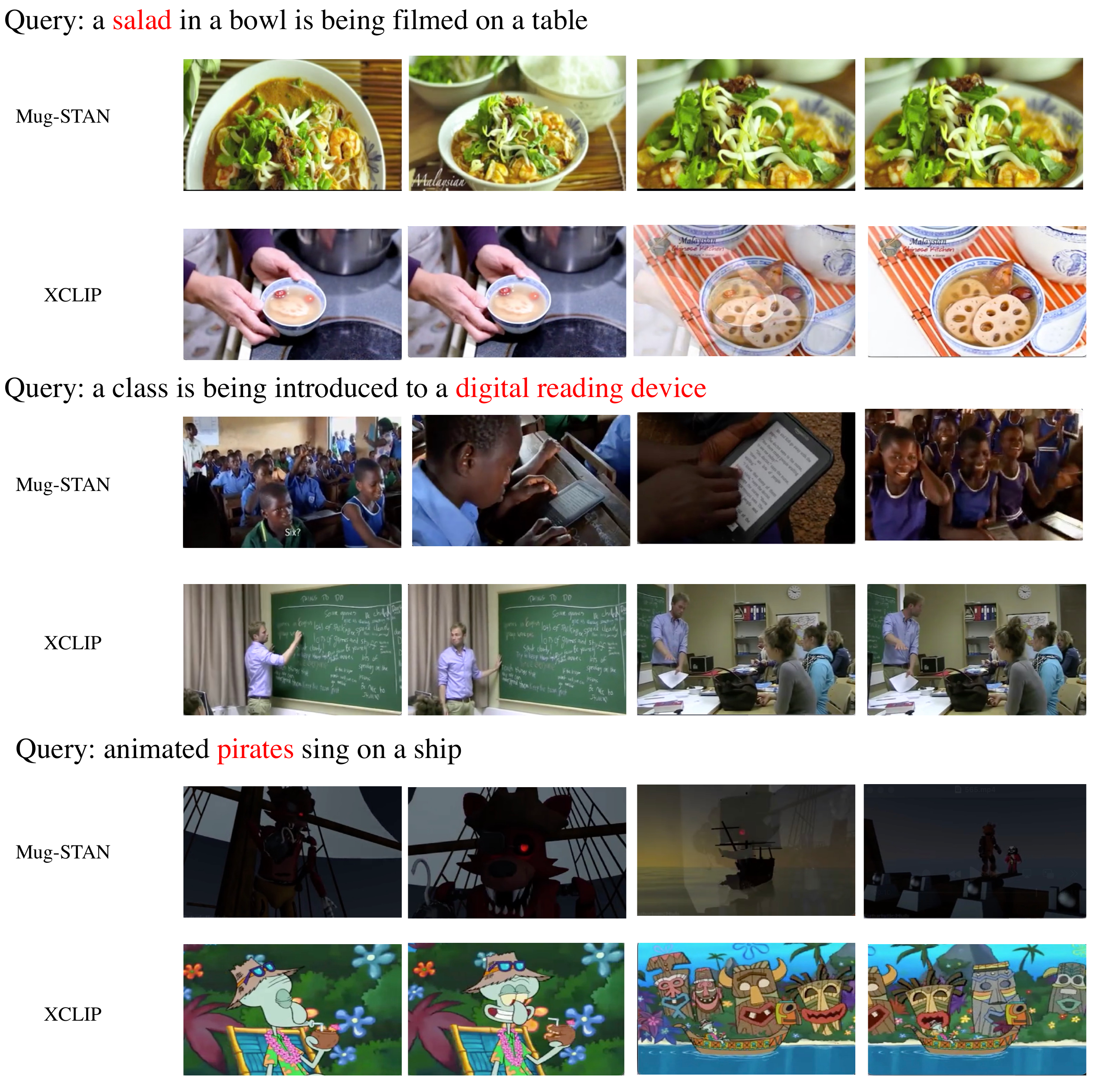} 
    \caption{Qualitative results of text-video retrieval on MSR-VTT. Given a text query, we present the correct matched video returned by Mug-STAN in the first row, and show the false result of XCLIP in the second row. The word highlighted in red indicates the key content missed in the false result.}
    \label{noun}
    \vspace{-0.6em}
\end{figure}

\begin{figure}[h] 
    \setlength{\abovecaptionskip}{-0.1 cm}
    \centering
    \includegraphics[width=1\columnwidth]{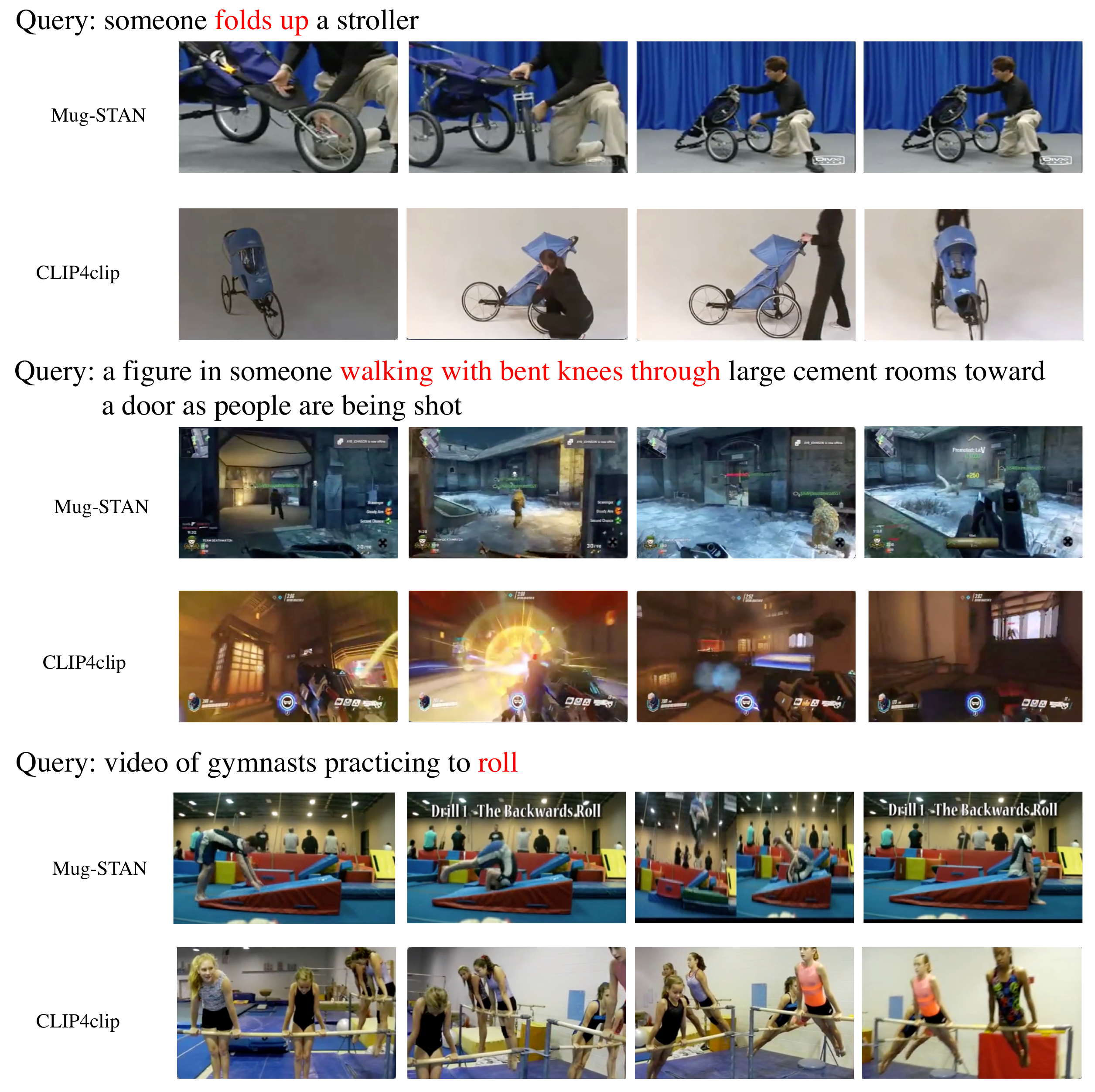} 
    \caption{Qualitative results of text-video retrieval on MSR-VTT. Given a text query, we present the correct matched video returned by Mug-STAN in the first row, and show the false result of CLIP4clip in the second row. The word highlighted in red indicates the key content missed in the false result.}
    \label{verb}
    \vspace{-0.6em}
\end{figure}
  
\textit{Can Mug-STAN work on image-language pretrained models beyond CLIP?} 
To verify the generalizability of our method, we further implement Mug-STAN based on another famous image-text pretrained model, \textit{i.e.,} CoCa\cite{yu2022coca}. We only use the visual and text encoder of CoCa and load the pretrained weights released by \textit{OpenCLIP}, which is pretrained on LAION2b \cite{schuhmann2022laion}.  As is illustrated in Table \ref{coca}, compared to the \textit{CoCa} baseline, which is directly fine-tuned on downstream tasks with mean pooling as its temporal modeling strategy, both STAN and Mug bring significant performance improvement, while the post-pretraining on WebVid10M further boost the finetuning result. The expermental results demonstrate that Mug-STAN has the potential to be migrated to various emergent image-text pretrained models.

\vspace{0.2em}

\begin{figure}[h] 
    \setlength{\abovecaptionskip}{-0.1 cm}
    \centering
    \includegraphics[width=1\columnwidth]{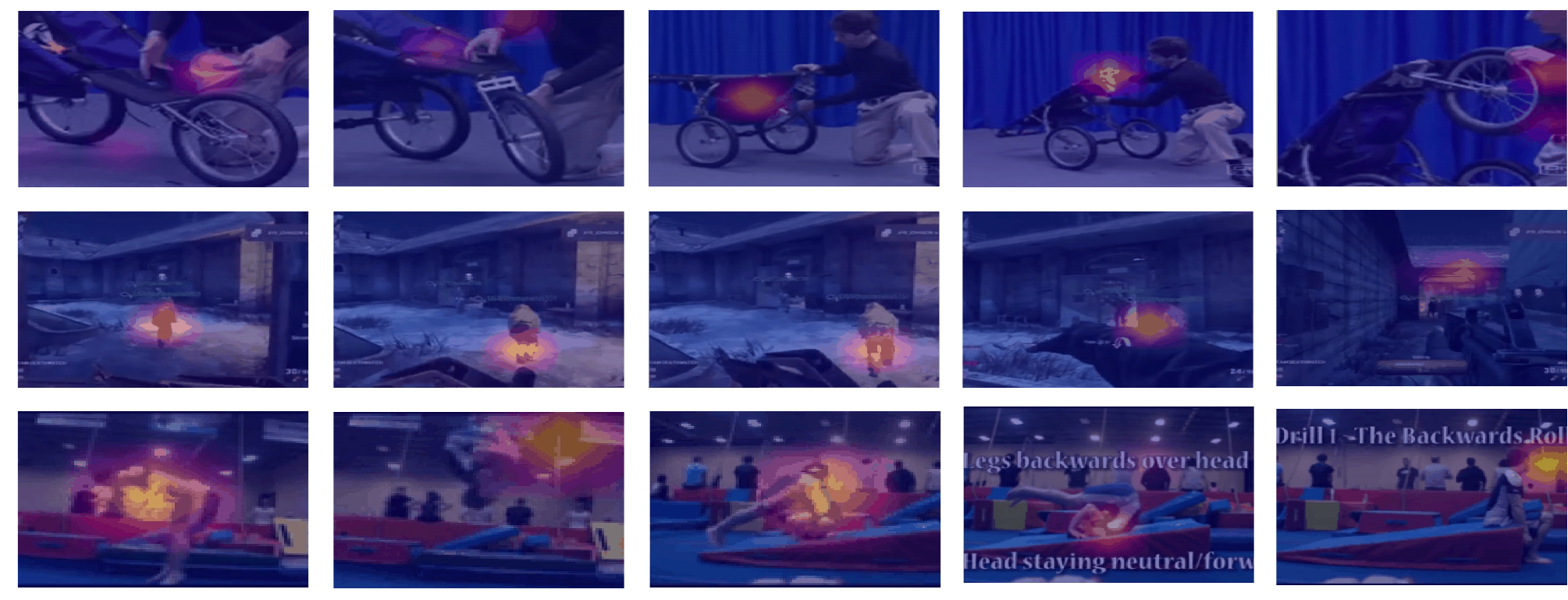} 
    \caption{Visualziation of intra-frame module of STAN on MSR-VTT. Given a text query. The region in red gains more attention from the model. We visualize the attention with VideoCAM.}
    \label{att}
    \vspace{-0.6em}
\end{figure}

\begin{figure}[h] 
    \setlength{\abovecaptionskip}{-0.1 cm}
    \centering
    \includegraphics[width=1\columnwidth]{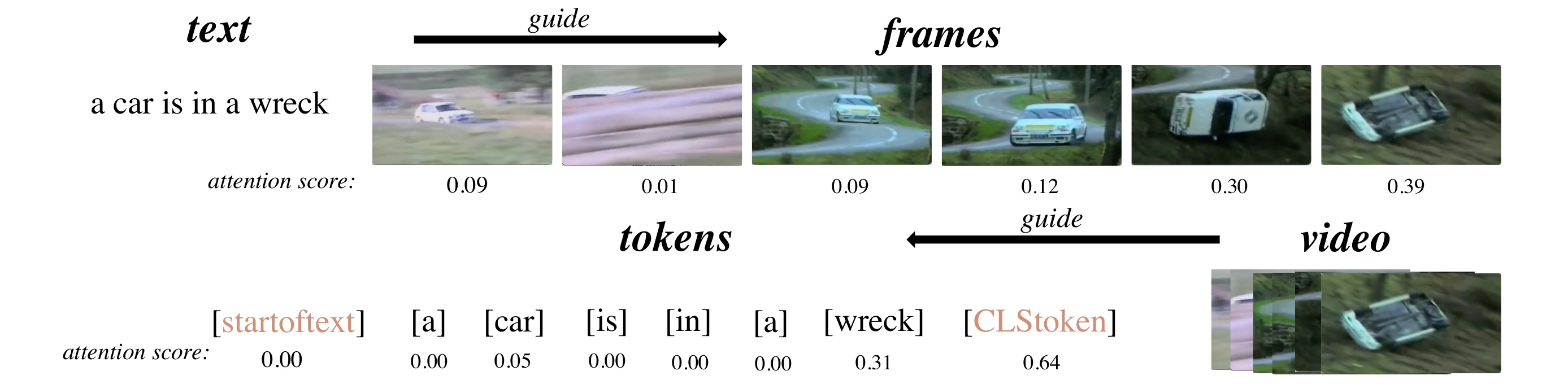} 
    \caption{The qualitative result of the softmax scores of sentence guiding frames in Eq .\ref{eq4} and video guiding tokens in Eq. \ref{eq8}.}
    \label{visual_mug}
    \vspace{-0.2em}
\end{figure}

\textit{What is the best hyper-parameter setting of STAN?}
STAN functions as a new branch positioned alongside the pretrained visual backbone, which takes the video frame representation at different levels of pretrained visual layers as inputs. To study impact of different setting of STAN, we present extensive ablation study for STAN-CLIP-B/32 in Fig. \ref{stan_ablation} on both video-language tasks and video-only tasks. The first is the number of STAN layers, as is shown, for MSRVTT retrieval, the performance enhancement of STAN reaches its peak at 4 layers, after which the performance begins to decline with further increases of layers; On SSv2, the performance improvement of STAN seems to stabilize after 6 layers. Overall, using STAN with 4 to 6 layers is recommended as a suitable choice for various tasks, considering the optimal balance between performance gains and computational efficiency. Secondly is the location of STAN layer. We fix the number of STAN layers to 4 and align STAN layers with 1-4, 5-8 and 9-12 CLIP layers respectively. The results suggest that the mid-to-high level of pretrained CLIP representation holds more significance for downstream tasks. Then, we align the last layer of CLIP and STAN, and vary the interval of selected CLIP layers between the STAN layers, \textit{e.g.,}, interval=2 means STAN receives outputs of the
6th, 8th, 10th, and 12th layers. As shown in Table, interval=1 is the best choice for both datasets. Finally is the number of the whole STAN networks. We find that introducing more STAN layers makes no difference on MSRVTT but can bring a slight improvement to ssv2, but this is not cost-effective considering the increase in computational complexity.

\vspace{0.2em}

\begin{figure}[h] 
    \setlength{\abovecaptionskip}{-0.1 cm}
    \centering
    \includegraphics[width=1\columnwidth]{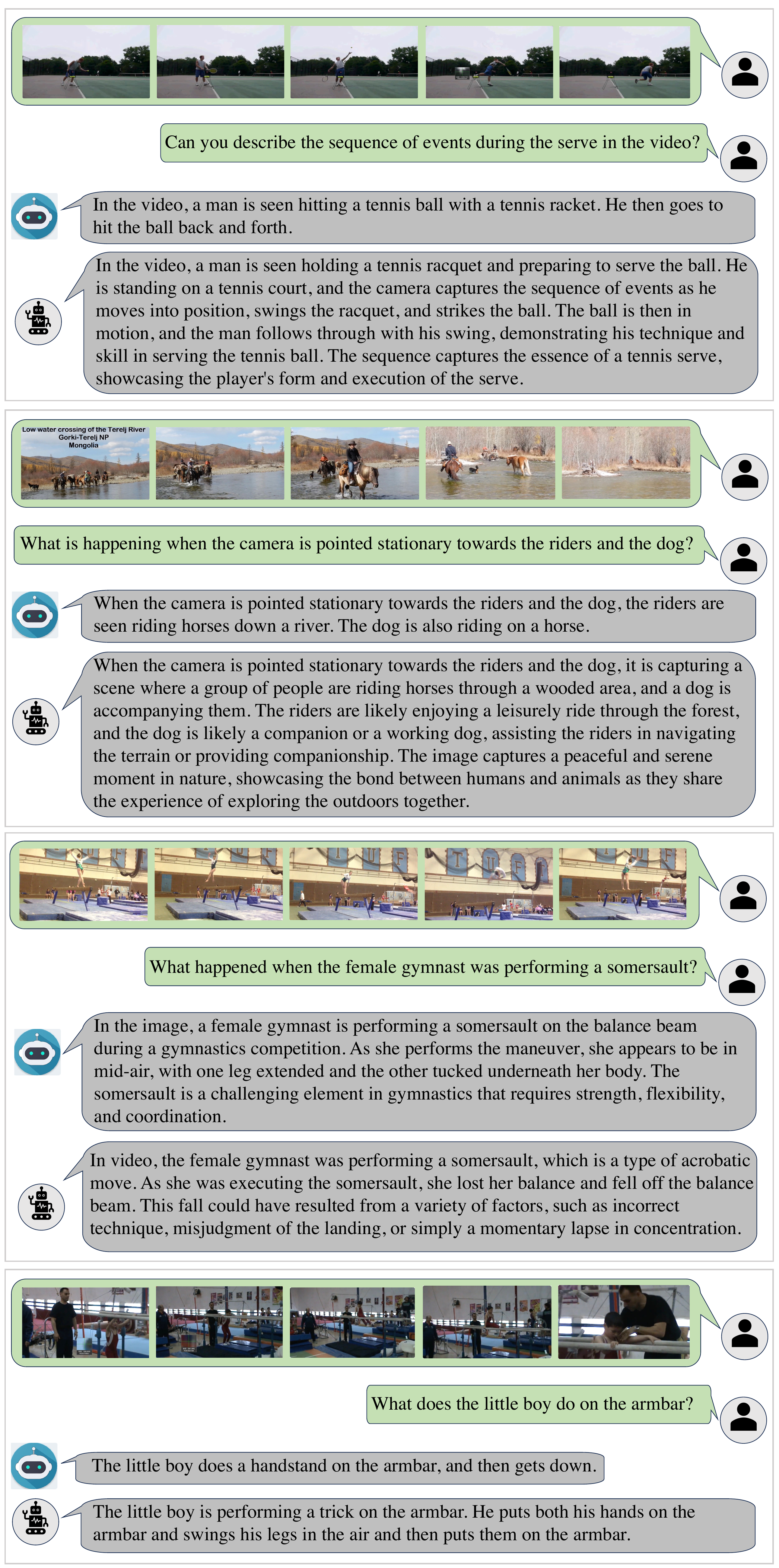} 
    \caption{The qualitative results of video chatting. We showcase the results from LLaVa (above) and STAN-LLaVa (below).}
    \label{chat}
    \vspace{-0.6em}
\end{figure}

\textit{Is Mug the optimal design for aligning videos and text?}
To understand the optimal design for the video-text interaction module, we perform a detailed ablation analysis. Initially, we explore the granularity of interaction within Mug. In Table \ref{ablation_mug} (middle), \textit{Frame-Text Interaction} indicates substituting \textit{video-guided text embedding} in Mug with conventional [CLS] token embedding, while \textit{Video-Token Interaction} represents substituting \textit{text-guided video embedding} with conventional averaged frame-wise embedding. The results demonstrate that \textit{text-guided video embedding} is more important than the \textit{video-guided text embedding}, which reveals that the partial-misalignment problem is more severe in the video modality. Then, we investigate different cross-modal interaction strategies. A well-known interaction modeling module is WTI in DRL \cite{wang2022disentangled} and its follower hunyuan \cite{jiang2022tencent}, which learns single-modality based attention scores to determine which token-frame scores are most representative of text-video correspondence. In contrast, Mug utilizes token-frame correspondence scores to introduce cross-modal mutual guidance, where the most relevant parts between the video-text pair, which potentially have higher scores, would be highlighted. Table \ref{ablation_mug} (bottom) shows that Mug outperforms WTI and hunyuan in terms of performance. Besides, the mutual-guided cross-modal embedding aggregation in Mug is akin to a soft key concept selection process. To explore this idea,   we further replace the softmax operations in Eq.\ref{eq3},\ref{eq4},\ref{eq7},\ref{eq8} with a top-k hard selection operation. However, we find that the optimal ``top-k'' value varies across datasets, and the module with softmax consistently outperforms the others on both datasets.

\textit{Is Mug effective for video-text misalignment?}
In Table \ref{ablation_misalign}, we investigate the effectiveness of Mug in addressing misalignment. Firstly, as depicted in Fig \ref{intro2}(b), video-text datasets often exhibit varying degrees of misalignment. In the upper bound, we further post results of Mug on data with different levels of misalignment. Notably, the majority of Mug's improvements are observed in cases of moderate to severe misalignment. The more pronounced the misalignment in video-text data, the greater the performance gains Mug exhibits over STAN. Furthermore, we conduct a comparison with other methods for mitigating misalignment, as indicated in the lower bound of the table. To ensure a fair comparison, following \cite{han2022temporal}, we pretrain Mug-STAN on HTM-370K and evaluate its zero-shot performance on datasets as presented in \cite{han2022temporal}. The results clearly demonstrate that Mug-STAN has a significant advantage over other state-of-the-art methods when operating under the same experimental conditions. In summary, our experimental findings across various dimensions consistently highlight the effectiveness of Mug in addressing misalignment in video-text data.

\subsection{Qualitative Results}
In experiments, we substantiated Mug-STAN's capacity for proficient temporal modeling, all the while harnessing the benefits of pretrained knowledge. Expanding on these quantitative findings, we now present qualitative results that unveil the efficacy of Mug-STAN across these two aspects.

First of all,  we showcase text-to-video retrieval outcomes of the intermediate-structure based method XCLIP \cite{ni2022expanding} and our Mug-STAN. Illustrated in Fig.\ref{noun}, these instances can be effortlessly resolved if a model can effectively align the emphasized object concepts in queries, like "salad," with videos that contain corresponding visual content. However, XCLIP produces inaccurate outcomes by returning results where the crucial objects are missing from the videos. This comparison underscores the limitation of the intermediate structure in effectively transferring high-level visual-text alignment knowledge, the work at which our method excels. 
Subsequently, we provide comparison results of text-to-video retrieval for CLIP4clip \cite{luo2022clip4clip} and Mug-STAN in Fig.\ref{verb}. The figure demonstrates that CLIP4clip, which is based on the posterior structure, produces incorrect outcomes. Although the results encompass accurate static contexts as described in queries (such as ``stroller" and ``gymnasts"), they feature erroneous dynamic information that doesn't align with the emphasized concepts in the queries (such as ``folds up" and ``roll"). These results emphasize that our approach can more effectively harness spatial-temporal information for enhanced video comprehension.
Then, we visualize the attention of Mug-STAN's intra-frame module using VideoCAM, as depicted in Fig.\ref{att}. These visualizations demonstrate that our STAN module consistently directs its attention towards critical content within videos, spanning across different moments in time. 
Finally, to shed more light to the effectiveness of Mug, we present a qualitative result of the cross-modal guidance. In Fig. \ref{visual_mug}, we present both the text-to-frame correspondence scores $\widetilde{s_i}$ (above) and video-to-token correspondence scores $\widetilde{s'_j}$ (below). The results show that for text-to-frame guidance, most of the attention is focused on the last two frames where the cars are rolling, which contain the most relevant information with the text. For video-to-token guidance, the attention is guided towards the tokens ``car'', ``wreck'', and the ending token (CLS token). It reveals that \textit{Mug} efficiently enhancing the aligned parts in the video-text pair for cross-modal alignment. 

\subsection{Video Chatting}
The domain of natural language processing has undergone a significant transformation with the introduction of pretrained Large Language Models (LLMs). 
The achievements of LLMs have also hastened the advancement of AI systems that integrate visual models with LLMs, enabling multimodal reasoning and action \cite{alayrac2022flamingo, li2023blip, liu2023visual, zhu2023minigpt}. Commonly, these models construct a projection from the output of the pretrained visual encoder (\textit{e.g.,} CLIP) to the input of the LLM. They then engage in visual instruction tuning, a process that facilitates multimodal interactions and conversations. 
Inspired by these visual-language chatbots, a new wave of methods has emerged that involve video chatting, which engages video backbone with LLMs and performs video instruction tuning \cite{li2023videochat}. Nonetheless, the training of video-language chatbots encounters similar challenges as video-language pretraining, namely huge computation costs and limited training source.

Fortunately, Mug-STAN offers a potential solution to these challenges. Unlike existing video chatbots, our approach does not involve the resource-intensive instruction tuning. Instead, we harness the power of existing image-language knowledge in a zero-shot manner. Specifically, we first post-pretraining Mug-STAN-CLIP on video-language datasets. Following this, we incorporate the pretrained branch networks into the visual backbone of image-language chatbots. Given that most existing multimodal chatting commonly utilizes a frozen CLIP as the visual backbone, our method can seamlessly empower image-language chatbots with the capacity for video understanding and processing. 
Lastly, the video token can also be seamlessly fed into the LLM for video chatting. This integration is facilitated by the fact that the output of STAN matches the token count of the image encoder. We take LLaVa \cite{liu2023visual} as pretrained image-text chatbots and present the qualitative results of STAN-LLaVa in Fig. \ref{chat}. 
Compared with LLaVa, our method empowers the chatbot to precisely narrate the events within the video sequence and accurately recognize temporal-extensive actions. Notably, these results are achieved without resorting to any instruction tuning. This underscores the significant potential of Mug-STAN in adapting pretrained image-language chatbots to the realm of videos. 

\section{Conclusion}
In this paper, we first investigate and identify the key point of adapting pretrained image-language models to video domains: building generalizable temporal modeling and suppressing video-text partial misalignment. To this end, we propose the novel Spatial-Temporal Auxiliary Network with Mutual-guided alignment module (Mug-STAN), where STAN utilizes the multi-level branch structure for effective temporal modeling and Mug introduces cross-modal mutual-guided feature aggregation to mitigate misalignment. Finally, we perform comprehensive experiments to demonstrate the superiority of Mug-STAN. Extensive experimental results show that our adaption method achieves state-of-the-art results on a broad range of video tasks. 

{
\bibliographystyle{IEEEtran}
\bibliography{ref}
}

\vfill

\end{document}